\documentclass{article} 
\usepackage{custom_tex_arxiv}

\title{Fair Classification by Direct Intervention\\ on Operating Characteristics}
\author{
  Kevin Jiang \quad
  Edgar Dobriban \\
  University of Pennsylvania\footnote{Author e-mail addresses: \texttt{kcjiang@wharton.upenn.edu}, \texttt{dobriban@wharton.upenn.edu}.}
}

\begin{document}

\maketitle

\begin{abstract}
We develop new classifiers under group fairness in the attribute-aware setting for binary classification with multiple group fairness constraints (e.g., demographic parity (DP), equalized odds (EO), and predictive parity (PP)). 
We propose a novel approach, applicable to linear fractional constraints, based on directly intervening on the operating characteristics of a pre-trained base classifier, by
(i) identifying optimal operating characteristics using the base classifier's group-wise ROC convex hulls and
(ii) post-processing the base classifier to match those targets.
As practical post-processors,
we consider randomizing a mixture of group-wise thresholding rules subject to minimizing the expected number of interventions. 
We further extend our approach to handle multiple protected attributes and multiple linear fractional constraints.
On standard datasets (COMPAS and ACSIncome), 
our methods simultaneously 
satisfy approximate DP, EO, and PP with few interventions and a near-oracle drop in accuracy; comparing favorably to previous methods.

\end{abstract}

\section{Introduction}
\label{sec:introduction}

Modern machine learning systems inherit and can amplify historical and measurement biases present in data, 
raising concerns in high-stakes applications such as criminal justice, credit, hiring, and healthcare \citep[see e.g.,][etc]{barocas2016big, obermeyer2019dissecting, raghavan2020mitigating, 10.1145/3376898,fuster2022predictably}. 
These concerns have galvanized the field of \emph{algorithmic fairness} \citep[see e.g.,][]{barocas2023fairness}, which formalizes fairness criteria and develops methods to enforce them.

One broad family of metrics, termed \emph{group fairness} criteria, requires that a performance metric of a model is equal across protected groups (e.g., race, gender, age).
These performance metrics are often motivated by legal rulings, regulatory guidance, and broader normative considerations (e.g., anti-discrimination norms, see \citet{hardt2016equality, verma2018fairness, barocas2023fairness}).

Methods designed to ensure group fairness can be roughly categorized into three categories based on where they intervene in the training pipeline: 
\emph{pre-processing} intervenes on the training data via methods like reweighing/sampling, causal inference, and adversarial debiasing \citep[see e.g.,][etc]{salazar2021fawos, zhang2020towards, zeng2024bayes, plevcko2024fairadapt};
\emph{in-processing} modifies the learner’s objective or model architecture \citep[see e.g.,][etc]{ZLM2018, agarwal2018reductions, cho2020fair}; 
and \emph{post-processing} alters the predictions of a fixed learned model \citep[see e.g.,][etc]{hardt2016equality, alghamdi2022beyond, xian2024unified}.
These methods have found broad applications \citep[see e.g.,][etc]{jammalamadaka2023responsible, Weerts_Fairlearn_Assessing_and_2023, mackin2025post}.

In parallel, policy and litigation 
have highlighted the desire to simultaneously satisfy multiple fairness constraints (e.g., analyses surrounding the COMPAS model, see \citet{LarsonAngwin2016ProPublicaTechResponse, DieterichMendozaBrennan2016NorthpointeResponse}). 
However, incompatibility results in the literature show that perfect compliance across sufficiently many metrics is generally impossible except in special cases \citep{chouldechova2017fair, kleinberg2018inherent, defrance2025maximal}.
Consequently, many works adopt \emph{approximate} group fairness, where a model's performance across groups are only required to be approximately equal (see, e.g., \citet{celis2019classification, zeng2024bayes}).

\paragraph{Our contributions.} In this work, we develop new classifiers aiming for approximate group fairness for binary classification with multiple group fairness constraints (e.g., demographic parity (DP), equalized odds (EO), and predictive parity (PP)), as long as they have a linear fractional representation related to that considered in \citet{celis2019classification}. 
Our novel approach directly intervenes on the operating characteristics of a pre-trained base classifier, by: 
(i) identifying optimal operating characteristics using the base classifier's group-wise receiver operating characteristic (ROC) convex hulls; and then
(ii) post-processing the base classifier to match those targets.
We extend our approach to handle multiple protected attributes and multiple linear fractional constraints.
On standard datasets (COMPAS and ACSIncome), 
our methods simultaneously 
satisfy approximate DP, EO, and PP.
They lead to few changes to the predictions made by the original classifier and a nearly optimal drop in accuracy; comparing favorably to previous methods.

\section{Related Work}
\label{sec:related_work}
Both empirically and theoretically, it is known that satisfying multiple fairness constraints can be challenging \citep{chouldechova2017fair,kleinberg2018inherent,majumder2023fair,defrance2025maximal}.
\citet{bell2023possibility} show that, in principle,
there may exist classifiers that achieve approximate fairness for multiple constraints.
However, this assumes the existence of classifiers that can satisfy an arbitrary assignment of labels to the test set, and also requires tuning population-level quantities that are typically out of our control.

In highly relevant work, 
\citet{celis2019classification} propose a meta-algorithm based on an optimization perspective in the space of classifiers.
Our approach differs fundamentally,
since we
rely on the geometry of the realizable regions traced by the convex hull of each group's ROC curve. 
\citet{hsu2022pushing} develop post-processing methods based on mixed-integer linear programming for a more granular notion of fairness, equalizing group rates across scores/predicted probabilities as opposed to binary labels. 
Their definitions reduce to ours when using two bins.
With this approach,
empirically, our methods outperform those from \citet{celis2019classification} and \citet{hsu2022pushing}.
Due to space limitations, additional related work is reviewed in Section \ref{arw}.

\section{Problem Setup}
\label{sec:setup}

We consider a fair binary classification problem where two types of feature are observed: the usual feature $X \in \mathcal{X}$ and the \emph{protected feature} $A \in \mathcal{A}$ which we take to be discrete, with $\mathcal{A} = \{1,2,\ldots,m\}$, $m\ge 2$.
We would like to achieve non-discrimination with respect to the \emph{protected groups}, defined by the values $a\in \mathcal{A}$ of the protected feature. The features and binary label are distributed according to $(X,A,Y)\sim \mathbb{P}$.
Given a pre-trained probabilistic predictor  $s:\mathcal{X}\times\mathcal{A}\to[0,1]$ and an i.i.d.\ sampled post-processing set\footnote{In the post-processing literature for fairness, this set has been alternatively called a \emph{holdout} or \emph{calibration} set \citep{pleiss2017fairness,hansen2024multicalibration}.} $\mathcal{D}_{\textrm{post}}:=\{(x_i,a_i,y_i)\}_{i=1}^{N}$ drawn from $\mathbb{P}$, we seek to develop a (possibly randomized) 
classifier $f:\mathcal{X}\times\mathcal{A}\to\{0,1\}$ that minimizes the population risk 
$ \mathcal{L}(f)=\E[\ell(f(X,A),Y)]$,
for a loss $\ell: \{0,1\} \times \mathcal{Y}\to\mathbb{R}$ subject to 
$K\ge 1$ \emph{fairness constraints}.

To impose fairness constraints,
 we consider equalizing certain 
 \textit{group performance functions} $G_{k,a}$,
 which reflect the performance of the classifier over protected groups. 
We focus on loss functions and group performance rates which can be expressed as ratios of linear combinations of a classifier's true positive rate (TPR) and false positive rate (FPR); as well as the associated false negative rate (FNR) and  true negative rate (TNR), for each group\footnote{
When it is clear which classifier $f$ the group-wise operating characteristics are tied to, 
we will sometimes omit the classifier from the argument, 
writing, for instance $\TPR_a$, etc.} $a\in\mathcal{A}$:
\begin{align}\label{eq:pop_operating_rates}
    \TPR_a(f):=\Pr(f{=}1\mid Y{=}1,A{=}a),&\qquad
    \FPR_a(f):=\Pr(f{=}1\mid Y{=}0,A{=}a); \nonumber
    \\
    \FNR_a(f):=1-\TPR_a(f),&\qquad \TNR_a(f):=1-\FPR_a(f).
\end{align}

We consider the following class of group performance functions, a special case of more general notions from \citet{celis2019classification}, 
but encompassing many common fairness metrics, see Table \ref{tab:LF-metrics}).

\begin{definition}[LF group performance functions]\label{def:LF-group-performance}
    For each group $a$ and constraint $k$, a linear fractional (LF) group performance function $G_{k,a}$ is defined as
    \begin{align*}
        G_{k,a}(f) := \frac{\langle \vec{u}_{k,a},\vec{\rho}_a\rangle}{\langle \vec{v}_{k,a},\vec{\rho}_a\rangle}
        \quad\text{with}\quad
        \vec{\rho}_a := (\TPR_a(f),\FPR_a(f),1)^\top,\langle \vec{v}_{k,a},\vec{\rho}_a\rangle>0,
    \end{align*}
    where $\vec{u}_{k,a},\vec{v}_{k,a}\in\mathbb{R}^3$ are coefficient vectors that may depend on the underlying population distribution, but not on the classifier $f$. 
    A linear group performance function $ G_{k,a}(f) = {\langle \vec{u}_{k,a},\vec{\rho}_a\rangle}$ is obtained when $\vec{v}_{k,a}=(0,0,1)^\top$ selects the constant component only.
\end{definition}
\textit{Linear fractional (linear) fairness constraints}
are functions of only linear fractional (linear) group performance functions.
For instance, 
demographic parity \citep{CKP2009} is a linear fairness constraint, since it requires $G_{\mathrm{DP},a}(f) - G_{\mathrm{DP},a'}(f)=0$ for all $a \neq a'$, where the group-wise selection rate is, with $\pi_a := \Pr(Y{=}1 \mid A{=}a)$,
    \begin{align*}
    G_{\mathrm{DP},a}(f)
    &= \Pr(f{=}1 \mid A{=}a)
    = \pi_a\TPR_a + (1-\pi_a)\FPR_a
    = \frac{\langle \vec{u}_{\mathrm{DP},a},\vec{\rho}_a\rangle}{\langle \vec{v}_{\mathrm{DP},a},\vec{\rho}_a\rangle},
    \end{align*}
    with $\vec{u}_{\mathrm{DP},a}=(\pi_a,1-\pi_a,0),\
    \vec{v}_{\mathrm{DP},a}=(0,0,1)$.
    \textit{Predictive parity} \citep{chouldechova2017fair} is a linear fractional constraint, since it requires $G_{\mathrm{PP},a}(f) - G_{\mathrm{PP},a'}(f)=0$ for all $a \neq a'$, where
    the positive predictive value is
    \begin{align*}
    G_{\mathrm{PP},a}(f)
    &= \Pr(Y{=}1 \mid f{=}1, A{=}a)
    = \frac{\pi_a\TPR_a}{\pi_a\TPR_a + (1-\pi_a)\FPR_a} 
    = \frac{\langle \vec{u}_{\mathrm{PP},a},\vec{\rho}_a\rangle}{\langle \vec{v}_{\mathrm{PP},a},\vec{\rho}_a\rangle},
    \end{align*}
    where $\vec{u}_{\mathrm{PP},a}=(\pi_a,0,0),\
    \vec{v}_{\mathrm{PP},a}=(\pi_a,1-\pi_a,0)$,
    with denominator $\langle \vec{v}_{\mathrm{PP},a},\vec{\rho}_a\rangle=\Pr(f{=}1\mid A{=}a)>0$ for groups with nonzero selection rate.

\begin{table}
\caption{ 
Fairness metrics covered by linear fractional group performance measures.}
\label{tab:LF-metrics}
\begin{center}
\renewcommand{\arraystretch}{1}

\begin{tabular}{lll}
\multicolumn{1}{c}{\bf metric} &
\multicolumn{1}{c}{\bf type} &
\multicolumn{1}{c}{\bf coefficients $(\vec{u}_{k,a};\vec{v}_{k,a})$} \\
\hline \\
Demographic parity (DP) 
& linear 
& $\vec{u}=(\pi_a,1-\pi_a,0);\vec{v}=(0,0,1)$
\\
Equal opportunity (TPR parity) 
& linear 
& $\vec{u}=(1,0,0);\vec{v}=(0,0,1)$
\\
Predictive equality (FPR parity) 
& linear 
& $\vec{u}=(0,1,0);\vec{v}=(0,0,1)$
\\
Equalized odds (TPR \& FPR parity) 
& linear pair
& --- 
\\
Predictive parity (PPV parity) 
& lin.–frac.
& $\vec{u}=(\pi_a,0,0);\vec{v}=(\pi_a,1-\pi_a,0)$
\\
False omission rate (FOR) parity 
& lin.–frac.
& $\vec{u}=(-\pi_a,0,\pi_a);\vec{v}=(-\pi_a,-(1-\pi_a),1)$
\\
Accuracy parity 
& linear 
& $\vec{u}=(\pi_a,-(1-\pi_a),1-\pi_a);\vec{v}=(0,0,1)$
\\
\end{tabular}
\end{center}
\footnotesize
\noindent\textit{Notes.} Here, $\pi_a=\Pr(Y{=}1\mid A{=}a)$ denotes the group prevalence/base rate. 
For linear–fractional (lin.–frac.) metrics, denominator positivity is required: for PPV, $\pi_a\TPR_a+(1-\pi_a)\FPR_a>0$ (nonzero selection); 
for FOR, $\pi_a(1-\TPR_a)+(1-\pi_a)(1-\FPR_a)>0$ (nonzero non-selection).
\end{table}

Moreover, since satisfying multiple fairness constraints exactly may be impossible \citep{chouldechova2017fair, kleinberg2018inherent, defrance2025maximal},
we consider 
\emph{$\vec\delta$-approximate fairness}, as e.g., \citep{xian2024unified, zeng2024bayes}.
This requires pairwise differences of group performance functions to differ by no more than $\delta_k \ge 0$:
$|G_{k,a}(f) - G_{k,a'}(f)|\leq \delta_k, ~\forall a\neq a', k\in[K],$
where we further collect all the user-specified disparities $\delta_k$ as a single vector $\vec{\delta}\in[0,1]^k$.

We consider 
loss functions  that are a linear combination of \emph{linear} group performance functions\footnote{The usual misclassification rate 
$
\mathcal{L}(f)
=
\sum_{a\in\mathcal{A}} p_a \left[\pi_a(1-\TPR_a) + (1-\pi_a)\FPR_a\right]
=
\Pr\big(f(X,A)\neq Y\big)$,
 can be recovered by taking coefficients $\vec{\gamma}_a$ to be
$\vec{\gamma}_a=\left(-p_a\pi_a, p_a(1-\pi_a),p_a\pi_a\right)$,
$p_a=\Pr(A{=}a), \pi_a=\Pr(Y{=}1\mid A{=}a)$.}
$\mathcal{L}(f)
=
\sum_{a\in\mathcal{A}} \left\langle \vec{\gamma}_a,\vec{\rho}_a \right\rangle$,
$\vec{\rho}_a = (\TPR_a,\FPR_a,1)^\top, \vec{\gamma}_a\in\mathbb{R}^3$.
Altogether, on the population level, our fairness constrained optimization problem is then
\begin{tcolorbox}[colback=white,colframe=black,title={Population-level Optimal Classification with Linear Fractional Fairness Constraints}]
\begin{align}\label{eq:function-opt-problem}
 &\min_{f:\mathcal{X}\times\mathcal{A}\to\{0,1\}}
\quad \sum_{a\in\mathcal{A}} \left\langle \vec{\gamma}_a,\vec{\rho}_a(f) \right\rangle\quad  
\text{s.t.}
 \max_{a,a'\in\mathcal{A}} \left| G_{k,a}(f) - G_{k,a'}(f) \right| \le \delta_k, k\in[K], \\
& G_{k,a}(f)= \frac{\langle \vec{u}_{k,a}, \vec{\rho}_a(f)\rangle}{\langle \vec{v}_{k,a}, \vec{\rho}_a(f)\rangle}, 
\langle \vec{v}_{k,a}, \vec{\rho}_a(f)\rangle > 0,
\forall a,k;\quad 
 \vec{\rho}_a(f) := \left(\TPR_a(f),\FPR_a(f),1\right)^\top, \nonumber
\end{align}
\end{tcolorbox}

where $\vec{\gamma}_a\in\mathbb{R}^3$ are fixed loss-coefficient vectors; $\vec{u}_{k,a},\vec{v}_{k,a}\in\mathbb{R}^3$ are fixed coefficient vectors for constraint $k\in[K]$ and group $a\in\mathcal{A}$; and $\delta_k\geq 0$ are prescribed fairness tolerances for each $k\in[K]$.

The Bayes optimal regression function $\eta_a$ takes values, for all $x$, 
$\eta_a(x) := \Pr(Y{=}1|X{=}x,A{=}a)$, $a\in\mathcal{A}$.    
We call a classifier $f:\mathcal{X}\times\mathcal{A}\to\{0,1\}$ a \emph{group-wise thresholding rule (GWTR)} 
if it thresholds $\eta_a$ at group-specific values $t_a$, $a\in \mathcal{A}$:
    \begin{align}\label{eq:GWTR}
    f(x,a)= 1\left(\eta_a(x) > t_a\right)\quad\textrm{for some}\quad t_a\in[0,1],a\in\mathcal{A}.
    \end{align}

In what follows, we will be considering the convex hull of ROC curves, in which case points on the hull are obtained by allowing a mixture of thresholding rules.
\begin{definition}[Mixed-GWTR]\label{def:mixed_gwtr}
    A (possibly randomized) classifier $f:\mathcal{X}\times\mathcal{A}\to\{0,1\}$ is a \emph{mixed-GWTR} if, for each group $a\in\mathcal{A}$, there is
    a finite index set $J_a$, thresholds $\{t_{a,j}\in[0,1]\}_{j\in J_a}$, and nonnegative weights $\{\lambda_{a,j}\}_{j\in J_a}$ with $\sum_{j\in J_a}\lambda_{a,j}=1$, such that
    \begin{equation}\label{eq:mixed_gwtr}
    \mathbb{P}(f(x,a)=1\mid x,a)=\sum_{j\in J_a} \lambda_{a,j} \cdot1\left(\eta_a(x) > t_{a,j}\right).
    \end{equation}
\end{definition}
For a broad family of fairness constraints defined by linear fractional
group performance functions, an optimal solution to the population problem \eqref{eq:function-opt-problem} can be realized by a GWTR with suitable thresholds $\{t_{a^*}\}_{a\in\mathcal{A}}$ \citep{celis2019classification}.
This result, however, is not directly practicable,
as it only describes the optimum at the level of the unknown population from which the data has been sampled,
and presumes access to the regression functions $\eta_a$.
In practice, one must estimate these functions
as well as $\{t_a\}_{a\in\mathcal{A}}$.
Due to this estimation step,
in challenging situations when the sample size is small, the feasibility constraints 
can possibly become infeasible. 
We will see experimentally that 
this step
can indeed be highly noise-sensitive. 

\section{Fair classification via ROC feasibility regions}\label{sec:ROCF}

Motivated by the above observations, 
our approach relies on the following two steps.

\medskip
\noindent\textbf{(A) Restricting to post-processors.}
Given
a pre-trained probabilistic predictor  $s:\mathcal{X}\times\mathcal{A}\to[0,1]$---intended to approximate the label probabilities---and a post-processing set $\mathcal{D}_{\textrm{post}}=\{(x_i,a_i,y_i)\}_{i=1}^N$, 
we consider \emph{post-processors}
obtained from $s$:
\begin{align}\label{eq:post-processor}
    \mathcal{F}_N \equiv \mathcal{F}_N(s;\mathcal{D}_{\textrm{post}}):=\{f(\cdot,a) \text{ obtained from } s(\cdot,a) \text{  and }\mathcal{D}_{\textrm{post}}
    ~\text{(and possibly randomized)}\}.
\end{align}

\medskip
\noindent\textbf{(B) Moving to operating characteristic space.}
Instead of optimizing over functions $f$, 
we directly work with their \emph{group-wise operating characteristics} (or, \emph{rates}).
For each group $a$, define the \emph{realizable (or post-processing) ROC region}
\begin{equation}\label{eq:post-processing-region}
\mathcal{R}_a(s):=\left\{(\tpr,\fpr)\Big|\exists f\in\mathcal{F}_N\text{such that}\left(\TPR_a(f),\FPR_a(f)\right)=(\tpr,\fpr)\right\}.
\end{equation}

    Allowing randomized thresholding as in Definition \ref{def:mixed_gwtr}
    ensures that the set of achievable rates starting from any given $s$ is convex; 
    so $\mathcal{R}_a(s)$ coincides with the convex hull of the ROC points generated by thresholding $s(\cdot,a)$.\footnote{Hence, to restrict to a post-processor, 
    we equivalently require the classifier's group rate vectors to lie in its ROC convex hull (see \S\ref{sec:region-search}).}

\paragraph{Fair classification via operating characteristic feasibility regions (ROCF).} 
Our method, ROCF, departs from traditional post-processing methods  (and associated direct threshold search)
by directly working with operating characteristics. 
We work with post-processors,
constraining each group’s rates to lie in the (population) realizable ROC region $\widetilde{\mathcal{R}}_a(s)$ 
(see \S\ref{sec:rate-space-reform} for the population setting and \S\ref{sec:region-search} for its empirical convex hull analog).

We next provide an overview of the steps.
Due to space considerations, 
some details are presented in the supplementary material.

\subsection{Rate-space reformulation \& Post-processing Condition}\label{sec:rate-space-reform}
Let $\vec{\rho}_a=(\TPR_a,\FPR_a,1)^\top$ be as before, and define the lifted realizable set 
$\widetilde{\mathcal{R}}_a(s) :=\{\rho=(\tpr,\fpr,1)^\top : (\tpr,\fpr) \in \mathcal{R}_a(s)\}$.
To incorporate the post-processing constraint, and to move to the low-dimensional space of operating characteristics, we reformulate  \eqref{eq:function-opt-problem} to the form:
\begin{tcolorbox}[colback=white,colframe=black,
  title={Population-level Optimal Fair Classification via Post-Processor Operating Characteristics }]
\[
\refstepcounter{equation}%
\vcenter{\hbox{$\begin{aligned}
&\min_{\{\vec{\rho}_a\}_{a\in\mathcal{A}}}\quad 
\sum_{a\in\mathcal{A}} \big\langle \vec{\gamma}_a,\vec{\rho}_a \big\rangle\qquad 
\text{s.t.}\quad 
\max_{a,a'\in\mathcal{A}} \left| G_{k,a}(\vec{\rho}_a) - G_{k,a'}(\vec{\rho}_{a'}) \right| \le \delta_k,
k\in[K]; \\
& G_{k,a}(\vec{\rho}_a)=\frac{\langle \vec{u}_{k,a},\vec{\rho}_a\rangle}{\langle \vec{v}_{k,a},\vec{\rho}_a\rangle},
\langle \vec{v}_{k,a},\vec{\rho}_a\rangle>0,\forall a,k; \quad
 \vec{\rho}_a\in \widetilde{\mathcal{R}}_a(s),\forall a\in\mathcal{A}.
\end{aligned}$}}%
\tag{\theequation}\label{eq:post-processing-function-opt-problem}
\]
\end{tcolorbox}

\subsection{Operating characteristic Feasibility Regions and Centroid-based linearization}\label{sec:centroid-linearization}

The objective from \eqref{eq:post-processing-function-opt-problem} can be non-convex in $\{\vec{\rho}_a\}_{a\in\mathcal{A}}$ due to the linear fractional group performance constraints. 
Following \citet{celis2019classification} and \citet{xian2024unified}, we note that ensuring bounded pairwise differences $ |G_{k,a} - G_{k,a'}| \le \delta_k $ is equivalent to the existence of a \emph{centroid} $ q_k \in [0,1]$ with $ |G_{k,a} - q_k| \le \delta_k/2 $ for all $a\in\mathcal{A}$.
Thus, to reduce the number of constraints, 
for each constraint $ k \in [K] $, introduce a centroid $ q_k $. 
 Moreover, since pure linear constraints stay linear after the introduction of the centroids, we split the constraints:
Let $ \mathcal{K}_{\textrm{L}} $ and $ \mathcal{K}_{\textrm{LF}} $ denote the indices of linear and linear fractional fairness constraints, respectively, so that $\mathcal{K}_{\textrm{L}} ~\sqcup~ \mathcal{K}_{\textrm{LF}} = [K]$.

\paragraph{Linear constraints via centroids.}
For $ k \in \mathcal{K}_{\textrm{L}} $, the disparity constraint in \eqref{eq:post-processing-function-opt-problem} is equivalent to the existence of a centroid $ q_k $ such that
$-\frac{\delta_k}{2} \le\langle \vec{u}_{k,a}, \vec{\rho}_a \rangle - q_k \le\frac{\delta_k}{2}, \forall a \in \mathcal{A}$.
These are linear inequalities in $ (\vec{\rho}_a, q_k) $, so we can treat $ q_k $ as a decision variable while still having an LP.

\paragraph{Linear--fractional constraints via fixed centroids.}
For $ k \in \mathcal{K}_{\textrm{LF}} $, write $ U_{k,a}(\rho) := \langle \vec{u}_{k,a}, \rho \rangle $ and $ V_{k,a}(\rho) := \langle \vec{v}_{k,a}, \rho \rangle $.
Assuming $ V_{k,a}(\vec{\rho}_a) > 0 $, the centroid form
$
\left|{U_{k,a}(\vec{\rho}_a)}/{V_{k,a}(\vec{\rho}_a)} - q_k \right| \le \tfrac{\delta_k}{2}
$
is equivalent  to the pair of linear inequalities
\begin{equation}\label{eq:lin-frac-centroid}
\begin{aligned}
    U_{k,a}(\vec{\rho}_a) - \left(q_k + \tfrac{\delta_k}{2}\right) V_{k,a}(\vec{\rho}_a) &\le 0,\quad
    \left(q_k - \tfrac{\delta_k}{2}\right) V_{k,a}(\vec{\rho}_a) - U_{k,a}(\vec{\rho}_a) \le 0,
\end{aligned}
~\forall a \in \mathcal{A}.
\end{equation}
Thus, for any fixed centroid $ q_k $, the linear fractional constraint becomes linear in $ \vec{\rho}_a $.
Given $\vec{q}=(q_k)_{k\in\mathcal{K}_{\mathrm{LF}}}$,
the set $\mathfrak{F}(\vec{q};s,\vec{\delta})$ of $\Big\{ \{\vec{\rho}_a\}_{a\in\mathcal{A}}$ satisfying the above constraints$\Big\}$ is called an
\emph{operating characteristic feasibility region} (see Definition \ref{def:feasibility-region}), which we simply refer to as \emph{feasibility region} for convenience.

\paragraph{Inner linear program for fixed linear fractional centroids.}
Given centroids $\{q_k\}_{k\in\textrm{LF}}$, we obtain the following linear optimization problem in the rate variables $\vec{\rho}_a$ and the linear-centroids $q_k$, with $\widetilde{\mathcal{R}}_a$ from the beginning of \Cref{sec:rate-space-reform} and 
$ V_{k,a}(\rho) = \langle \vec{v}_{k,a}, \rho \rangle $ from above \eqref{eq:lin-frac-centroid}:
\begin{tcolorbox}[colback=white,colframe=black,
  title={Inner Optimization for Fixed Linear Fractional Centroids}]
\begin{equation}\label{eq:linear-opt-prob}
\begin{aligned}
& \min_{\{\vec{\rho}_a\}_{a\in\mathcal{A}},\{q_k\}_{k\in\mathcal{K}_{\textrm{L}}}} \quad
\sum_{a\in\mathcal{A}} \left\langle \vec{\gamma}_a, \vec{\rho}_a \right\rangle, 
 \text{s.t.}
-\tfrac{\delta_k}{2} \le \langle \vec{u}_{k,a}, \vec{\rho}_a \rangle - q_k \le \tfrac{\delta_k}{2},
\forall a \in \mathcal{A},k \in \mathcal{K}_{\textrm{L}}, \\
&  
\text{\eqref{eq:lin-frac-centroid} holds}~\&~ V_{k,a}(\vec{\rho}_a) >0, ~\forall k \in \mathcal{K}_{\textrm{LF}};\quad
\vec{\rho}_a \in \widetilde{\mathcal{R}}_a(s), \forall a \in \mathcal{A}.
\end{aligned}
\end{equation}
\end{tcolorbox}

Moreover, at the population level $ \widetilde{\mathcal{R}}_a$ is convex. 
If it is represented polyhedrally (as will be the case empirically via convex hulls in \S\ref{sec:region-search}), then \eqref{eq:linear-opt-prob} is a linear program.
In practice, for stability purposes, we additionally enforce that the denominators 
are strictly bounded away from zero via the linear constraints
$V_{k,a}(\vec{\rho}_a) \ge\varepsilon_k > 0,
    \forall a \in \mathcal{A}, k \in \mathcal{K}_{\textrm{LF}}$,
for small positive constants $ \varepsilon_k $ that we set later.

\paragraph{Outer search over linear fractional centroids.}
For each linear fractional constraint $k\in\mathcal{K}_{\mathrm{LF}}$, we restrict the centroid to a compact interval $ \mathcal{Q}_k \subset (0,1) $ that is consistent with post-processing and denominator positivity, i.e., choices of $q_k$ for which \eqref{eq:lin-frac-centroid} holds.
We show later that we can take $ \mathcal{Q}_k = [\delta_k/2,1-\delta_k/2]$.
We call these \emph{admissible} centroids.

We then search over $ \vec{q} \in \mathcal{Q} := \prod_{k\in\mathcal{K}_{\mathrm{LF}}} \mathcal{Q}_k $ (e.g., via a coarse grid search) and solve the inner linear program (cf. \eqref{eq:linear-opt-prob}) at each $\vec{q}$.
The explicit construction of $ \mathcal{Q}_k $ and the exact linear bands for common LF metrics (e.g., predictive parity) are deferred to the supplementary material (see \S\ref{app:imp:centroids}).

This outer search for fixed values of linear fractional centroids is justified by the following theorem (proof is deferred to Appendix \ref{app:proofs}).
\begin{theorem}\label{thm:centroid}
    Define the value function $\Phi$ such that 
    $\Phi(\vec{q})$ is the optimal objective value of \eqref{eq:linear-opt-prob} for any $\vec{q} \in \mathbb{R}^{|\mathcal{K}_{\mathrm{LF}}|}$.
    Then, the value of the
    optimization problem from \eqref{eq:post-processing-function-opt-problem} is equal to
    $\min_{\vec{q} \in \mathcal{Q}} \Phi(\vec{q})$.
    Moreover, 
    we can find an optimizer $\{\vec{\rho}_a^\star\}_{a\in\mathcal{A}}$
    of the objective in \eqref{eq:post-processing-function-opt-problem} 
    by selecting any minimizer $\vec{q}^\star\in\arg\min_{\vec{q}\in\mathcal{Q}}\Phi(\vec{q})$ and 
then optimizing the objective in \eqref{eq:linear-opt-prob} at $\vec{q}^\star$.
\end{theorem}


{\bf Empirical region search over ROC-hull supports.}
Given the post-processing set,
we search over plug-in ROC-hull supports and solve modest-sized LPs for centroid-specific feasibility. A feasibility guard ensures that, when needed,
tolerances are minimally relaxed (see \S\ref{sec:region-search} for details).

\subsection{Constructing classifiers}\label{sec:construct-classifiers}

We now seek to construct classifiers that achieve the
optimal rates.
Since there are many classifiers that can achieve any operating characteristic, the practitioner can impose additional desiderata such as minimizing the expected number of interventions, which we define as label flips from the base classifier $f^{(0)}$ (see \S\ref{sec:interv}).
Here, we propose two versions of an approach for this problem.

\subsubsection{Randomization procedure}\label{sec:randomization-procedure}
Take the base post-processor $f^{(0)}$ to be a mixed-GWTR (see \eqref{eq:mixed_gwtr}) whose operating point lies on the boundary of the convex hull of the empirical ROC curve, i.e., for all $x,a$:
\begin{align}\label{eq:baseline-postprocessor}
    f^{(0)}(x,a)=(1-\theta_a)1 \left(s(x,a)\ge t_{h,a}\right) +
    \theta_a 1\left(s(x,a)\ge t_{h+1,a}\right),\qquad 
    \theta_a\in[0,1],
\end{align}
where $t_{h+1,a}\ge t_{h,a}$, $~t_{h,a}, t_{h+1,a}\in\widehat{\mathcal{H}}_a$ are two adjacent points on the support of the empirical convex hull\footnote{Since $\widehat{\mathcal{H}}_a$ is constructed solely from a probabilistic predictor  and a post-processing set, and we add randomization implicit in the mixing parameter $\theta_a$, $f^{(0)}$ is indeed a valid post-processor (cf. \eqref{eq:post-processor}).} 
 (cf. \S\ref{sec:region-search}).
Let, for all $x,a$, $q_a(x):=\Pr(f^{(0)}=1\mid X=x,A=a)$ be the positive prediction rate of the base post-processor, and denote the operating characteristics of $f^{(0)}$ as 
${\TPR}^{(0)}_a:=\Pr(f^{(0)}{=}1\mid Y{=}1,A{=}a),
{\FPR}^{(0)}_a:=\Pr(f^{(0)}{=}1\mid Y{=}0,A{=}a),
\FNR^{(0)}=1-\TPR^{(0)}, \TNR^{(0)}=1-\FPR^{(0)}.
$
The group-wise optimal operating characteristics are fixed at values $(\widetilde{\TPR}_a, \widetilde{\FPR}_a)$, e.g., outputs from Algorithm \ref{alg:region-search-feasibility-guard}. 

Then the final post-processor $\widetilde{f}$ randomizes $f^{(0)}$
aiming to ``regularize" or ``shrink" $q_a$ towards a classifier that does not depend on the inputs $x$. 
This transformation also shifts the operating characteristics of $f^{(0)}$ 
towards the non-random classifier. 
There are several possible ways to implement this transformation, and here we discuss two. 

The \texttt{AntiDiagonal} method 
directly randomizes between $f^{(0)}$ and a fully random classifier $f^{(r)}(x,a)\sim\mathrm{Bernoulli}(p_a)$ with mixture weight $\lambda_a\in[0,1]$.
The final prediction is distributed as a Bernoulli random variable with success probability
\begin{equation}\label{eq:antidiagonal-form}
    \Pr\big(\widetilde{f}(x,a)=1\mid X=x,A=a\big)
    =(1-\lambda_a)q_a(x)+\lambda_a p_a.
\end{equation}

Its operating characteristics are thus the convex combination
\begin{equation}\label{eq:antidiagonal-rates}
    {\widetilde\TPR}_a=(1-\lambda_a){\TPR}^{(0)}_a+\lambda_a p_a,
    \qquad
    {\widetilde\FPR}_a=(1-\lambda_a){\FPR}^{(0)}_a+\lambda_a p_a.
\end{equation}
Geometrically, \texttt{AntiDiagonal} traces a line segment from the baseline operating point to the diagonal point $(\tpr,\fpr)=(p_a,p_a)$, or equivalently, an anti-diagonal point $(\FNR,\FPR)=(1-p_a,p_a)$.
Fixing the baseline post-processor and the optimal operating points then uniquely determines the values of $(\lambda_a,p_a)$.

Since $f^{(0)}$ is a mixed-GWTR as in \eqref{eq:baseline-postprocessor}, its operating point along any hull edge
can be parameterized by a single mixing parameter $\theta_a\in[0,1]$ between adjacent supports $(h,h{+}1)$:
\begin{equation}\label{eq:edge-param}
    \FNR^{(0)}_{a,\theta}=(1-\theta_a)\FNR_{a}^{(h)}+\theta_a\FNR_{a}^{(h+1)},
    \FPR^{(0)}_{a,\theta}=(1-\theta_a)\FPR_{a}^{(h)}+\theta_a\FPR_{a}^{(h+1)}.
\end{equation}
Given a target operating characteristic $\left(\widetilde{\FNR}_a,\widetilde{\FPR}_a\right)$ and fixed mixing parameter $\theta_a$, the parameters of randomization are uniquely determined by the rate-matching equations.
\footnote{Although we focused on fairness constraints expressible as pairwise differences in LF/L functionals, our procedure for constructing classifiers provides recipes to match any given target operating characteristics that lie in the convex hull of the empirical ROC curve of the base probabilistic predictor $s$, subject to minimizing the expected number of interventions (Algorithm~\ref{alg:minintervention}). }
Using 
\eqref{eq:antidiagonal-rates} with the baseline $\left(\FNR^{(0)}_{a,\theta},\FPR^{(0)}_{a,\theta}\right)$, we obtain
\begin{equation}\label{eq:antidiagonal-closed-form}
    \lambda_{a,\theta}=\frac{\widetilde{\FPR}_a+\widetilde{\FNR}_a-\left(\FPR^{(0)}_{a,\theta}+\FNR^{(0)}_{a,\theta}\right)}{1-\left(\FPR^{(0)}_{a,\theta}+\FNR^{(0)}_{a,\theta}\right)},
    p_{a,\theta}=\frac{\widetilde{\FPR}_a-\left(1-\lambda_{a,\theta}\right)\FPR^{(0)}_{a,\theta}}{\lambda_{a,\theta}}.
\end{equation}

Feasibility requires $0\le\lambda_{a,\theta}\le 1$ and $0\le p_{a,\theta}\le 1$. When $\lambda_{a,\theta}=0$, the target operating point coincides with the baseline operating point. The denominator in $\lambda_{a,\theta}$ vanishes when $\FPR^{(0)}_{a,\theta}+\FNR^{(0)}_{a,\theta}=1$, which only occurs for post-processors with degenerate ROC curves.

\texttt{LabelFlipping} is a mathematically equivalent approach, which
flips labels with outcome-dependent probabilities. Due to space limitations, we refer to Section \ref{lf} for the details.
We present the details of minimizing interventions with these two approaches in Section \ref{sec:minintervention-problem}. 
Empirically, these procedures have comparable performance (see \S\ref{sec:empirical}).

\begin{remark}
    While  \citet{hardt2016equality} considered label flipping
    and threshold search in some special cases, they did these separately; which limits the set of operating points their classifier can attain.    
    {More recently, \cite{hsu2022pushing} used label flipping randomization 
    on top of GWTRs to simultaneously satisfy multiple fairness constraints. 
    However, their procedure implicitly fixes the group-wise thresholds at the medians of the group-specific probabilistic predictor scores.
    These thresholds are overly restrictive, as the set of operating characteristics attainable from label flipping may not include the optimal operating point (see \S\ref{sec:exp:primary-result}). 
    }
\end{remark}

{\bf Multiple protected attributes.} Our method directly supports multiple protected attributes.
See Section \ref{sec:exp:multiple-protected-attributes} for the details.

\section{Experimental Results}
\label{sec:empirical}

We evaluate our method on standard empirical datasets,  aiming to simultaneously satisfy demographic parity (DP); equality of opportunity/TPR parity (EOpp), see \cite{hardt2016equality}; predictive equality/FPR parity (PEq), see \cite{CSFG2017}; predictive parity (PP), see \cite{chouldechova2017fair}; and possibly false omission rate parity (FOR-parity), see \cite{barocas2023fairness}.\footnote{The first three are linear constraints, while the latter two are LF constraints (see \S\ref{sec:setup}).}
while maximizing accuracy. 
Enforcing both $\delta_{\textrm{EOpp}}$ and $\delta_{\textrm{PEq}}$-approximate fairness is equivalent to enforcing $\max\{\delta_{\textrm{EOpp}}, \delta_{\textrm{PEq}}\}=\delta_{\textrm{EO}}$-approximate fairness for equalized odds (EO), see \cite{hardt2016equality}. 

Due to space constraints, we defer full implementation details, training hyperparameters, and exact baseline settings to the appendix \S\ref{app:imp}.
We use the \texttt{COMPAS} \citep{LarsonAngwin2016ProPublicaTechResponse} and \texttt{ACSIncome} \citep{ding2021retiring} datasets, with a TRAIN/POST/TEST$=30/35/35$ split: TRAIN fits $s$ (using a three-layer neural net), and POST fits all post-processors.
We report accuracy and five fairness metrics (DP, EOpp, PEq, PP, FOR-parity), 
aggregating mean results over 50 random seeds, along with standard deviations. 

\paragraph{Baselines.}
We compare with
two post-processing methods that seek to simultaneously control LF fairness constraints (\texttt{META} \citep{celis2019classification} and \texttt{MFOpt} \citep{hsu2022pushing}), 
and one state-of-the-art post-processing method that controls for linear fairness constraints (\texttt{LPP}, \citep{xian2024unified}). In addition, we record the performance of the unconstrained probabilistic classifier $s$ (\texttt{Baseline}) and an infeasible oracle post-processor that maximizes accuracy subject to the fairness constraints using the TEST set (\texttt{Oracle}), as an upper bound \citep{bell2023possibility}.

\subsection{Results}\label{sec:exp:results}

We present the primary results here in the main text and leave additional experimental results to the appendix (see Tables~\ref{tab:compas-acsincome-2} and \ref{tab:compas-acsincome-multiple-LF-constraints-2} in \S\ref{app:add-exp:results}).

\subsubsection{Approximate fairness for DP, EOpp, PEq, and PP}\label{sec:exp:primary-result}
We impose \textit{four} constraints (DP, EOpp, PEq, and PP) simultaneously, controlling for $\vec\delta$-approximate fairness at the level $\delta_i{=}0.05$ for all $i$.
Results are shown in Table \ref{tab:compas-acsincome}, where 
our methods \texttt{ROCF-AD/LF} refer to the \texttt{AntiDiagonal} and \texttt{LabelFlipping} modalities of Algorithm \ref{alg:minintervention}.

{\bf Results.}
First, we observe that our methods control all four disparity metrics. 
Among the baselines, only \texttt{META} achieves this on the \texttt{COMPAS} dataset, and no other baseline achieves it on \texttt{ACSIncome}. 
However, \texttt{META} exhibits a much larger accuracy drop than our methods. 
The reason for the performance gap might be that they aim to optimize in the space of classifiers, whereas our method achieves an advantage by working directly with the operating characteristics. 
Moreover, on both examples, we observe that our methods have accuracy close to the oracle.

{\bf Inevitable trade-offs.}
 Since even the oracle method exhibits an accuracy drop, 
 this suggests that fairness/accuracy trade-offs when imposing all four constraints might be unavoidable in our examples. 
 This is in contrast with simpler settings where we 
 only impose fewer constraints \citet{xian2024unified,celis2019classification, baumann2022enforcing}.

\begin{table}
\footnotesize
\caption{ Performance on the test set for (A) COMPAS ($|\mathcal{A}|{=}2$) and (B) ACSIncome ($|\mathcal{A}|{=}5$). 
The disparities $\delta_{\textrm{DP}}, \delta_{\textrm{EOpp}}, \delta_{\textrm{PEq}}, \delta_{\textrm{PP}}$ are controlled at level $0.05$ whenever they are active. 
\textbf{Interv.} is the empirical intervention rate on the test set (see Definition~\ref{def:intervention}). 
Cells in green indicate that the fairness constraint is satisfied within two standard deviations at level $0.05$, whereas cells in red indicate violation. 
Entries in the  Oracle rows are shaded in lighter colors to denote that they are not practically feasible baselines. }
\label{tab:compas-acsincome} 
\begin{center} 
\begin{tabular}{lccccccc}
\toprule
\multicolumn{8}{c}{\bf (A) COMPAS} \\
\addlinespace[0.35em]
\multicolumn{1}{c}{\bf Method} & \multicolumn{1}{c}{\bf Acc} & \multicolumn{1}{c}{\bf DP} & \multicolumn{1}{c}{\bf EOpp} & \multicolumn{1}{c}{\bf PEq} & \multicolumn{1}{c}{\bf PP} & \multicolumn{1}{c}{\bf FOR} & \multicolumn{1}{c}{\bf Interv.} \\
\midrule
Baseline & 0.68 $\pm$ 0.01 & \rcell{0.28 $\pm$ 0.05} & \rcell{0.27 $\pm$ 0.05} & \rcell{0.19 $\pm$ 0.05} & \gcell{0.07 $\pm$ 0.04} & 0.03 $\pm$ 0.02& 0.00 $\pm$ 0.00 \\
\addlinespace[0.25em]\cdashline{1-8}[0.4pt/1pt]\addlinespace[0.15em]
Oracle & 0.62 $\pm$ 0.02 & \gcelloracle{0.04 $\pm$ 0.01} & \gcelloracle{0.03 $\pm$ 0.02} & \gcelloracle{0.05 $\pm$ 0.01} & \gcelloracle{0.05 $\pm$ 0.00} & 0.16 $\pm$ 0.02 & N/A \\
\textbf{ROCF-AD} (ours) & 0.61 $\pm$ 0.02 & \gcell{0.05 $\pm$ 0.03} & \gcell{0.03 $\pm$ 0.02} & \gcell{0.05 $\pm$ 0.03} & \gcell{0.07 $\pm$ 0.04} & 0.15 $\pm$ 0.02 & 0.06 $\pm$ 0.02 \\
\textbf{ROCF-LF} (ours) & 0.61 $\pm$ 0.02 & \gcell{0.05 $\pm$ 0.03} & \gcell{0.03 $\pm$ 0.03} & \gcell{0.05 $\pm$ 0.03} & \gcell{0.07 $\pm$ 0.04} & 0.15 $\pm$ 0.02 & 0.06 $\pm$ 0.03 \\
\addlinespace[0.25em]\cdashline{1-8}[0.4pt/1pt]\addlinespace[0.15em]
MFOpt & 0.63 $\pm$ 0.01 & \rcell{0.26 $\pm$ 0.04} & \rcell{0.25 $\pm$ 0.04} & \rcell{0.21 $\pm$ 0.04} & \gcell{0.08 $\pm$ 0.04} & 0.07 $\pm$ 0.03 & 0.13 $\pm$ 0.02 \\
META & 0.50 $\pm$ 0.02 & \gcell{0.05 $\pm$ 0.17} & \gcell{0.05 $\pm$ 0.16} & \gcell{0.04 $\pm$ 0.18} & \gcell{0.06 $\pm$ 0.19} & 0.15 $\pm$ 0.05 & 0.00 $\pm$ 0.00 \\
\addlinespace[0.25em]\cdashline{1-8}[0.4pt/1pt]\addlinespace[0.15em]
LPP-DP & 0.67 $\pm$ 0.01 & \gcell{0.06 $\pm$ 0.03} & \gcell{0.04 $\pm$ 0.03} & \gcell{0.03 $\pm$ 0.03} & \rcell{0.15 $\pm$ 0.03} & 0.09 $\pm$ 0.03 & 0.00 $\pm$ 0.00 \\
LPP-EO & 0.67 $\pm$ 0.01 & \gcell{0.10 $\pm$ 0.04} & \gcell{0.09 $\pm$ 0.04} & \gcell{0.04 $\pm$ 0.03} & \rcell{0.14 $\pm$ 0.03} & 0.08 $\pm$ 0.03 & 0.00 $\pm$ 0.00 \\
\bottomrule
\end{tabular}


\begin{tabular}{lccccccc}
\toprule
\multicolumn{8}{c}{\bf (B) ACSIncome ($|\mathcal{A}|{=}5$)} \\
\addlinespace[0.35em]
\multicolumn{1}{c}{\bf Method} & \multicolumn{1}{c}{\bf Acc} & \multicolumn{1}{c}{\bf DP} & \multicolumn{1}{c}{\bf EOpp} & \multicolumn{1}{c}{\bf PEq} & \multicolumn{1}{c}{\bf PP} & \multicolumn{1}{c}{\bf FOR} & \multicolumn{1}{c}{\bf Interv.} \\
\midrule
Baseline & 0.79 $\pm$ 0.00 & \rcell{0.25 $\pm$ 0.01} & \rcell{0.24 $\pm$ 0.02} & \gcell{0.09 $\pm$ 0.02} & \rcell{0.23 $\pm$ 0.02} & {0.08 $\pm$ 0.01} & 0.00 $\pm$ 0.00 \\
\addlinespace[0.25em]\cdashline{1-8}[0.4pt/1pt]\addlinespace[0.15em]
Oracle & 0.69 $\pm$ 0.01 & \gcelloracle{0.05 $\pm$ 0.00} & \gcelloracle{0.05 $\pm$ 0.00} & \gcelloracle{0.03 $\pm$ 0.01} & \gcelloracle{0.05 $\pm$ 0.00} & {0.23 $\pm$ 0.01} & N/A \\
\textbf{ROCF-AD} (ours) & 0.69 $\pm$ 0.01 & \gcell{0.05 $\pm$ 0.01} & \gcell{0.05 $\pm$ 0.01} & \gcell{0.03 $\pm$ 0.01} & \gcell{0.07 $\pm$ 0.03} & {0.23 $\pm$ 0.01} & 0.03 $\pm$ 0.01 \\
\textbf{ROCF-LF} (ours) & 0.69 $\pm$ 0.01 & \gcell{0.05 $\pm$ 0.01} & \gcell{0.05 $\pm$ 0.01} & \gcell{0.03 $\pm$ 0.01} & \gcell{0.07 $\pm$ 0.03} & {0.23 $\pm$ 0.01} & 0.03 $\pm$ 0.01 \\
\addlinespace[0.25em]\cdashline{1-8}[0.4pt/1pt]\addlinespace[0.15em]
LPP-DP & 0.78 $\pm$ 0.00 & \gcell{0.06 $\pm$ 0.01} & \gcell{0.07 $\pm$ 0.01} & \rcell{0.09 $\pm$ 0.01} & \rcell{0.35 $\pm$ 0.01} & {0.15 $\pm$ 0.01} & 0.00 $\pm$ 0.00 \\
LPP-EO & 0.78 $\pm$ 0.00 & \rcell{0.12 $\pm$ 0.01} & \gcell{0.06 $\pm$ 0.02} & \gcell{0.06 $\pm$ 0.01} & \rcell{0.33 $\pm$ 0.01} & {0.14 $\pm$ 0.01} & 0.00 $\pm$ 0.00 \\
\bottomrule
\end{tabular}
\end{center}
\end{table}

{\bf Interventions.}
We also observe that the number of interventions is small, around $6\%$ for \texttt{COMPAS} and $3\%$ for \texttt{ACSIncome} for both of our methods. 

{\bf Scalability.}
Our method scales well to the much larger \texttt{ACSIncome} dataset, demonstrating both practical computational runtimes (see \S\ref{app:add-exp:runtimes}) and attaining the nominal disparity levels across all fairness constraints.
The latter is to be expected, since our method can better approximate the population level feasibility region 
and the realizable ROC region $\mathcal{R}_a(s)$ (cf. \eqref{eq:post-processing-region}) with a better trained probabilistic predictor  and a larger post-processing set.

\subsubsection{Two linear fractional constraints: adding false omission rate parity}\label{sec:exp:multiple-LF-constraints}

{\bf  False omission rate parity.}
We consider satisfying an additional linear fractional constraint, 
the \emph{false omission rate parity} (FOR-parity) along with predictive parity and  equality of opportunity constraint.
Including the false omission rate is important because it captures disparities in the reliability of negative predictions, ensuring that groups are not disproportionately subjected to undetected errors when they are predicted as negative.
We increase the tolerance level to $\vec{\delta}{=}0.10$ and exclude demographic parity and predictive equality 
in this experiment, 
since we often find that the empirical feasibility region $\widehat{\mathfrak{F}}(s,\vec{\delta};\mathcal{D}_{\textrm{post}})$ (see Remark \ref{remark:empirical-feasibility-region}) is empty when controlling at a $\vec{\delta}{=}0.05$ level.
Indeed, additionally satisfying a low level of FOR-parity is substantively nontrivial:
\citet{majumder2023fair} show through extensive empirical studies that fairness metrics cluster differently, with FOR-parity separating from DP/EO and PP.

{\bf Methods compared.}
Existing baselines do not explicitly support this configuration out of the box: \citet{hsu2022pushing} only handles $\vec\delta$-approximate DP, EO, and PP, and, while \citet{celis2019classification} can, in principle, encode any combination of LF/L fairness constraints, their released code does not implement FOR-parity \citep{celis2019-fairclassification-code}. 
Consequently, we report results only for our method (\texttt{ROCF-AD/LF}), the optimal operating point (\texttt{Oracle}), and the linear post-processing method of \citet{xian2024unified} that only controls for equality of opportunity (\texttt{LPP-EOpp}).
 The latter is included merely as a strong recent baseline from the literature, and we emphasize that it does not aim to ensure predictive parity and false omission rate parity. 

{\bf Results.}
 Table~\ref{tab:compas-acsincome-multiple-LF-constraints} reports results for the \textsc{ACSIncome} dataset with multiple protected attributes.
 As we can see, our methods control all three fairness constraints at the desired level, while achieving a near-oracle accuracy. 
 This reinforces the effectiveness of our method.
 In contrast, the \texttt{LPP-EOpp} method does not control predictive parity; which is reasonable as it does not aim to do so.  
Moreover, we observe a substantial accuracy-fairness tradeoff of $21$ percent. 
This suggests that satisfying fairness across multiple protected attributes for multiple LF constraints is difficult (Table \ref{tab:compas-acsincome-multiple-LF-constraints}).

\begin{table}
\footnotesize
\caption{ Performance on the test set for ACSIncome ($|\mathcal{A}|{=}5$). 
The same protocol as in Table \ref{tab:compas-acsincome}  is used, except
 $\delta_{\textrm{EOpp}}, \delta_{\textrm{PP}}$ and $\delta_{\textrm{FOR}}$ are controlled at level $0.10$. }
\label{tab:compas-acsincome-multiple-LF-constraints}
\begin{center} 
\begin{tabular}{lccccccc}
\toprule
\multicolumn{8}{c}{\bf ACSIncome ($|\mathcal{A}|{=}5$)} \\
\addlinespace[0.35em]
\multicolumn{1}{c}{\bf Method} & \multicolumn{1}{c}{\bf Acc} & \multicolumn{1}{c}{\bf DP} & \multicolumn{1}{c}{\bf EOpp} & \multicolumn{1}{c}{\bf PEq} & \multicolumn{1}{c}{\bf PP} & \multicolumn{1}{c}{\bf FOR} & \multicolumn{1}{c}{\bf Interv.} \\
\midrule
Baseline & 0.78 $\pm$ 0.01 & 0.24 $\pm$ 0.01 & \rcell{0.25 $\pm$ 0.03} & 0.09 $\pm$ 0.02 & \rcell{0.23 $\pm$ 0.03} & \gcell{0.08 $\pm$ 0.01} & 0.00 $\pm$ 0.00 \\
\addlinespace[0.25em]\cdashline{1-8}[0.4pt/1pt]\addlinespace[0.15em]
Oracle & 0.57 ± 0.04 & 0.48 ± 0.05 & \gcelloracle{0.10 ± 0.01} & 0.57 ± 0.08 & \gcelloracle{0.10 ± 0.01} & \gcelloracle{0.10 ± 0.01} & N/A \\
\textbf{ROCF-AD} (ours) & 0.57 ± 0.02 & 0.48 ± 0.03 & \gcell{0.11 ± 0.01} & 0.57 ± 0.05 & \gcell{0.11 ± 0.01} & \gcell{0.11 ± 0.03} & 0.06 ± 0.01 \\
\textbf{ROCF-LF} (ours) & 0.56 ± 0.02 & 0.49 ± 0.02 & \gcell{0.11 ± 0.01} & 0.59 ± 0.04 & \gcell{0.11 ± 0.01} & \gcell{0.11 ± 0.03} & 0.05 ± 0.01 \\
\addlinespace[0.25em]\cdashline{1-8}[0.4pt/1pt]\addlinespace[0.15em]
LPP-EOpp & 0.79 $\pm$ 0.00 & 0.15 $\pm$ 0.01 & \gcell{0.11 $\pm$ 0.01} & 0.06 $\pm$ 0.01 & \rcell{0.31 $\pm$ 0.01} & \gcell{0.12 $\pm$ 0.01} & 0.00 $\pm$ 0.00 \\
\bottomrule
\end{tabular}
\end{center}
\end{table}

\section{Discussion}\label{sec:discussion}

In this paper, we proposed an approach for fair classification with linear fractional approximate fairness constraints, which relied on reformulating the goal in the space of the operating characteristics of the classifiers and then designing fairness interventions at that level. 
We observed that our method compares favorably to existing baselines in experiments. 
An important direction for future work is that the proposed randomization approaches may
randomize any individuals,
but it could be of interest to 
design intervention policies that restrict randomization to only a subpopulation. 


\section*{Acknowledgments}

This work was partially supported by the NSF and the Sloan Foundation.

\bibliography{ref}
\bibliographystyle{iclr2026_conference}

\appendix
\section{Appendix}\label{app}

\subsection{Additional Related Work}
\label{arw}

Recent work by \citet{zehlike2025beyond} 
interpolates only between accuracy equality, false negative and false positive rate parity, leveraging optimal transport. 
Theoretical work by \citet{hamman2024unified} studies the tradeoff between demographic parity, equalized odds, and predictive parity using partial information decomposition by fixing one of the fairness metrics and exploring the tradeoffs between the other two, 
without explicitly designing classifiers.
Intervening on ROC curves has also been used to ensure forms of fairness different from the ones that we consider in this paper \citep{vummintala2025froc}.

Beyond our focus on post-processing methods, there are also pre- and in-processing methods that seek to modify the training process or dataset. 
Some of these aim to simultaneously control a broad set
of disparity measures by removing the dependence of the protected attribute $A$ on $(X,Y)$ \citep{ZLM2018, bothmann2023causal, plevcko2024fairadapt, leininger2025overcoming}. 

Post-processing is attractive when retraining is impractical or constrained by governance, cost, or latency—conditions that have become increasingly common with generative AI \citep{hu2022lora, dettmers2023qlora, wang2025parameter} and with deployed models in regulated domains (e.g., credit, hiring, healthcare; see \citet{quell2021machine, groves2024auditing, dupreez2025use}).

\subsection{Additional Methodological Details}

\subsubsection{Empirical region search over ROC-hull supports}\label{sec:region-search}

We instantiate the empirical analog of \eqref{eq:post-processing-function-opt-problem} by searching over linear fractional centroids for the best post-processing operating points from each group's empirical ROC convex hull.
More formally, let the plug-in estimates of the operating characteristics in \eqref{eq:pop_operating_rates} on the postprocessing set $\mathcal{D}_{\textrm{post}}$ be
\begin{align}\label{eq:empirical-operating-rates}
n_{a,1} = \sum_{i=1}^N 1\left(a_i=a,y_i=1\right), \qquad
n_{a,0} = \sum_{i=1}^N 1\left(a_i=a,y_i=0\right), \qquad n_a=n_{a,1}+n_{a,0}, \nonumber \\
\widehat{\TPR}_a(f)
= \frac{1}{n_{a,1}} \sum_{i=1}^N 1\left(a_i=a,y_i=1\right)\cdot1\left(f(x_i,a_i)=1\right), \nonumber \\
\widehat{\FPR}_a(f)
= \frac{1}{n_{a,0}} \sum_{i=1}^N 1\left(a_i=a,y_i=0\right)\cdot1\left(f(x_i,a_i)=1\right), \nonumber \\
\
\widehat{\FNR}_a(f) = 1-\widehat{\TPR}_a(f), \qquad
\widehat{\TNR}_a(f) = 1-\widehat{\FPR}_a(f).
\end{align}

Then, for each $a\in\mathcal{A}$, we form the group-wise empirical ROC $\{(\widehat{\TPR}_a^{(j)},\widehat{\FPR}_a^{(j)})\}_{j\in[n_a]}$
from the probabilistic predictor scores $s(x_i,a_i)$ and labels $y_i$ on the post-processing set by setting 
\begin{align*}
f^{(j)}(x,a)
&:= 1\left( \operatorname{rank}_a\left(s(x,a)\right)\le j\right), \qquad j\in[n_a] 
\end{align*}
in \eqref{eq:empirical-operating-rates} where\footnote{Scores are assumed to be continuous, so that a tie in the ranking does not occur; however, ties can be broken by a deterministic rule (e.g., via lexicographic ordering) so that exactly $j$ group-$a$ points satisfy $f^{(j)}(x_i,a_i)=1$.}
$\operatorname{rank}_a\left(s(x,a)\right)
:= 1+\sum_{i=1}^N 1\left(a_i=a, s(x_i,a_i)>s(x,a)\right).
$
We then retain the upper convex hull vertices
$\widehat{\mathcal{H}}_a=\{(\widehat{\TPR}_a^{(j)},\widehat{\FPR}_a^{(j)})\}_{j\in S_a}$ for some index set $S_a \subseteq [n_a]$ (e.g., using Andrew's monotone chain algorithm \citep{andrew1979another, o1998computational}), 
and
construct lifted operating characteristic vectors $\widehat{r}_a^{(j)}=(\widehat{\TPR}_a^{(j)},\widehat{\FPR}_a^{(j)},1)^\top$ from $\widehat{\mathcal{H}}_a$ and empirical plug-in coefficients for 
$\widehat{\gamma}_a,\widehat{u}_{k,a},\widehat{v}_{k,a}$ (e.g., replace $\pi_a$ by $\widehat{\pi}_a$).

Since any operating characteristic in the convex hull of $\widehat{\mathcal{H}}_a$ can be written as a linear combination of the convex hull support points, we enforce post-processing by introducing convex weights $\{\lambda_{a,j}\}_{j\in S_a}$ that satisfy
\begin{equation}\label{eq:rho-lambda-coupling}
    \sum_{j\in S_a}\lambda_{a,j}=1,\quad \lambda_{a,j}\ge 0,\qquad
    \widehat{\rho}_a = \sum_{j\in S_a}\lambda_{a,j}\cdot\widehat{r}_a^{(j)},
\end{equation}

where $\widehat{\rho}_a$ are the (lifted) operating points $(\widehat{\TPR}_a,\widehat{\FPR}_a,1)^\top$. The resulting optimization problem and full procedure for finding the optimal operating points is presented in Algorithm~\ref{alg:region-search} (\texttt{RegionSearch}).

\begin{remark}\label{remark:empirical-feasibility-region}
    {The region of operating points specified by the constraints of the inner LP (i.e., the linear fractional/linear fairness constraints and post-processing constraint) in Algorithm~\ref{alg:region-search} is the empirical analog of the centroid-specific population-level feasibility region $\mathfrak{F}(\vec{q}; s, \vec{\delta})$ discussed in \S\ref{sec:ROCF}. We call the union of this centroid specific region over the set of admissible centroids (cf. \S\ref{sec:centroid-linearization}) the
    \emph{empirical feasibility region} $\widehat{\mathfrak{F}}(s, \vec{\delta}; \mathcal{D}_{\textrm{post}})$.}
\end{remark}
\begin{remark}\label{remark:region-search-size-LP}
    {Each inner problem is a modest-sized LP with
    $\sum_{a\in\mathcal{A}} |S_a| + |\mathcal{K}_{\mathrm{L}}|$ decision variables. The number of linear constraints is often smaller:
     $|\mathcal{A}|+1$ for the simplex and post-processing per group, $2|\mathcal{A}||\mathcal{K}_{\mathrm{L}}|$ for the linear centroid bands, $2|\mathcal{A}||\mathcal{K}_{\mathrm{LF}}|$ for the LF bands, and $|\mathcal{A}||\mathcal{K}_{\mathrm{LF}}|$ for the denominator checks.}

    {Since the hull support sizes $|S_a|$ are typically small \citep[see e.g.,]{groeneboom2021grenander}, each LP solve is fast in practice, so that the overall runtime is dominated by the low-dimensional outer search over linear fractional centroids. We report computational runtimes in \S\ref{app:add-exp:runtimes}.}
\end{remark}

\paragraph{A feasibility guard.} To ensure that we produce a classifier even when \texttt{RegionSearch} does not yield a feasible operating point, we wrap a simple $\alpha$-expansion policy around the \texttt{RegionSearch} procedure of Algorithm \ref{alg:region-search}.

On a high level, if Algorithm~\ref{alg:region-search}  is infeasible with the initial, user-specified tolerances, 
we re-solve Algorithm \ref{alg:region-search} with fairness tolerances that are uniformly relaxed by an expansion factor of $\alpha$. 
By performing a bisection search on an appropriately designed interval for $\alpha$, our \texttt{RegionSearch} is guaranteed to produce a feasible operating point, from which we can construct a classifier (per \S\ref{sec:construct-classifiers}).

More formally, given user-specified tolerances $\vec{\delta}{=}\{\delta_k\}_{k\in\mathcal{K}_{\mathrm{L}}\cup\mathcal{K}_{\mathrm{LF}}}$, define
$
\vec{\delta}(\alpha):=\{\alpha\delta_k\}_{k\in\mathcal{K}_{\mathrm{L}}\cup\mathcal{K}_{\mathrm{LF}}}$ for $\alpha\ge 1$.
Let the unconstrained/baseline per-group operating points be, for $a\in\mathcal{A}$,
$j_a^\star \in\arg\min_{j\in S_a}\left\langle \widehat{\gamma}_a,\widehat{r}_a^{(j)}\right\rangle$
and
$\widehat{\rho}^{\mathrm{base}}_a:=\widehat{r}_a^{(j_a^\star)}$.
These are 
the solution of \texttt{RegionSearch} with no fairness constraints.\footnote{The objective is linear in the operating characteristics, so an optimum $\widehat{\rho}^{\mathrm{base}}_a$ occurs at a hull vertex.}
The corresponding baseline group performance functions are then
\begin{align*}
    g_{\ell,a}^{\mathrm{base}}:=\left\langle \widehat{u}_{\ell,a},\widehat{\rho}^{\mathrm{base}}_a\right\rangle, \qquad
    r_{k,a}^{\mathrm{base}}:=\frac{\left\langle \widehat{u}_{k,a},\widehat{\rho}^{\mathrm{base}}_a\right\rangle}{\big\langle \widehat{v}_{k,a},\widehat{\rho}^{\mathrm{base}}_a\big\rangle}
    \quad\text{(whenever }\langle \widehat{v}_{k,a},\widehat{\rho}^{\mathrm{base}}_a\rangle\ge \varepsilon_k\text{)},
\end{align*}

with baseline disparity measures 
$
    \Delta_\ell:=\max_{a} g_{\ell,a}^{\mathrm{base}}-\min_{a} g_{\ell,a}^{\mathrm{base}}$; 
    $\Delta_k :=\max_{a} r_{k,a}^{\mathrm{base}}-\min_{a} r_{k,a}^{\mathrm{base}}$.

Now, note that we can recover the baseline operating points $\widehat{\rho}^{\mathrm{base}}_a$ by re-solving \texttt{RegionSearch} with $\vec{\delta}(\alpha_{\textrm{hi}})$, where
$\alpha_{\textrm{hi}}$ is the maximum ratio between the baseline disparity measures and the nominal disparity levels. That is, an upper bound for the expansion parameter is 
$
\alpha_{\mathrm{hi}} := \max\left\{1,\max_{i\in\mathcal K_{\mathrm L}\cup\mathcal K_{\mathrm{LF}}}\tfrac{\Delta_i}{\delta_i}\right\}.
$

By design, $\{\widehat{\rho}^{\mathrm{base}}_a\}_{a\in\mathcal{A}}$ is feasible at $\alpha=\alpha_{\mathrm{hi}}$, and feasibility is monotone in $\alpha$: if the inner LP of \texttt{RegionSearch} is feasible at $\alpha$, it remains feasible for any $\alpha'\ge \alpha$. We therefore bisect on $\alpha\in[\alpha_{\textrm{lo}}, \alpha_{\textrm{hi}}] = [1,\alpha_{\mathrm{hi}}]$ by setting, at each step, $\alpha\leftarrow(\alpha_{\mathrm{lo}}+\alpha_{\mathrm{hi}})/2$, re-solving Algorithm~\ref{alg:region-search} with $\vec{\delta}(\alpha)$, and updating the search interval appropriately. 
For completeness, the full wrapper algorithm is provided in Algorithm \ref{alg:region-search-feasibility-guard} (\texttt{RegionSearch-FG}).

This feasibility guard produces a classifier while minimally relaxing the user-specified tolerances.
In our experiments, 
it rarely activates;
when it does, the expansion parameter is relatively small (e.g., $\alpha{\approx}1.04$ and see \S\ref{app:add-exp:results}).

This guard also provides practitioners a certificate of feasibility 
and, when feasibility is not met at the nominal tolerance levels, a minimal relaxation of the tolerances for which a post-processor (cf. \eqref{eq:post-processor}) can be constructed. 
Developing a more nuanced tradeoff
between fairness constraints 
(e.g., loosening $\delta_i$ while holding $\delta_j$ constant for $j{\neq}i$) is beyond the scope of the work, as these tradeoffs are application-dependent.

\subsection{Algorithms}\label{app:algs}
 In this section, we provide the pseudo-code for several of our proposed algorithms. We summarize each of the algorithms here for clarity.
\begin{itemize}
    \item \texttt{ROCF-Pipeline} (Algorithm~\ref{alg:rocf-pipeline}): Top level procedure that first finds target operating characteristics using a pre-trained classifier $s$, post-processing set $\mathcal{D}_{\textrm{post}}$, and user-specified tolerances $\vec{\delta}$ for LF/F fairness constraints (Algorithm~\ref{alg:region-search-feasibility-guard}). 
    Then, it
    constructs a classifier to achieve those operating characteristics with the option of imposing additional desiderata (e.g., minimizing expected number of interventions; see Algorithms~\ref{alg:construct-classifier} and~\ref{alg:minintervention}).
    
    \item \texttt{RegionSearch} (Algorithm~\ref{alg:region-search}): Empirical instantiation of the population level optimal fair classification problem (cf.~\eqref{eq:post-processing-function-opt-problem}); searches over each group’s empirical ROC-hull support to select target operating characteristics. Handles LF constraints via a small outer search over admissible centroids (cf.~\S\ref{app:imp:centroids}). 
    
    \item \texttt{RegionSearch-FG} (Algorithm~\ref{alg:region-search-feasibility-guard}): Wraps a feasibility guard around \texttt{RegionSearch} that  guarantees feasibility by loosening the tolerance levels $\vec{\delta}$ when the initial search is infeasible. 
    
    \item \texttt{ConstructClassifier} (Algorithm~\ref{alg:construct-classifier}): Constructs a randomized post-processor with
    target operating characteristics, subject to additional desiderata.
    
    \item \texttt{MinIntervention} (Algorithm~\ref{alg:minintervention}): Specific instantiation of Algorithm~\ref{alg:construct-classifier}.
    Given target operating characteristics, this algorithm produces a mixed-GWTR (cf. Definition~\ref{def:mixed_gwtr}) with added randomization,
    to achieve the target characteristics while minimizing the expected intervention rate (cf. Definition~\ref{def:intervention}). 
    Has two modalities (\textsc{AntiDiagonal} and \textsc{LabelFlipping}) corresponding to randomization procedures discussed in the text. 
\end{itemize}

\begin{algorithm}
\caption{ (End-to-end ROCF pipeline)}\label{alg:rocf-pipeline}
\begin{algorithmic}[1]
\Require Training set $\mathcal{D}_{\mathrm{train}}$, postprocessing set $\mathcal{D}_{\mathrm{post}}$, test set $\mathcal{D}_{\mathrm{test}}$; groups $\mathcal{A}$; 
tolerances $\{\delta_k\}$; constraint indices $\mathcal{K}_{\mathrm{L}},\mathcal{K}_{\mathrm{LF}}$~with $\mathcal{K}_{\mathrm{L}}\sqcup \mathcal{K}_{\mathrm{LF}} = [K]$; 
centroid intervals $\{\mathcal{Q}_k\}_{k\in\mathcal{K}_{\mathrm{LF}}}$; margins $\{\varepsilon_k>0\}$.


\State \textbf{Pre-train predictor.} Fit a probabilistic predictor $s:\mathcal{X}\times\mathcal{A}\to[0,1]$ on $\mathcal{D}_{\mathrm{train}}$.

\State \textbf{Postprocess ROC curve.} On $\mathcal{D}_{\mathrm{post}}$, compute probabilistic predictor scores $s(x_i,a_i)$ and retrieve labels $y_i$. For each $a\in\mathcal{A}$: form the empirical ROC curve and keep the upper convex hull supports $\widehat{\mathcal{H}}_a$; build lifted points $\widehat{r}_a^{(j)}=(\widehat{\TPR}_a^{(j)},\widehat{\FPR}_a^{(j)},1)^\top$. Form plug-in coefficients $\widehat{\gamma}_a,\widehat{u}_{k,a},\widehat{v}_{k,a}$.

\State \textbf{Search region (Algorithm~\ref{alg:region-search-feasibility-guard}).} Provide $\{\widehat{\mathcal{H}}_a\},\{\widehat{r}_a^{(j)}\},\{\widehat{\gamma}_a,\widehat{u}_{k,a},\widehat{v}_{k,a}\},\{\delta_k\},\mathcal{K}_{\mathrm{L}},\mathcal{K}_{\mathrm{LF}},\{\mathcal{Q}_k\},$ 
$\{\varepsilon_k\},\{\tau_\alpha\}$ to Algorithm~\ref{alg:region-search-feasibility-guard}; obtain target operating points $\{\widetilde{\rho}_a\}$.

\State \textbf{Construct classifier (Algorithm~\ref{alg:construct-classifier}).} Using $\{\widetilde{\rho}_a\}$ and $s$, construct a classifier $\widehat{f}$ that attains $\widetilde{\rho}_a$ for each group (optionally with additional desiderata).

\State \textbf{Evaluate.} On $\mathcal{D}_{\mathrm{test}}$, evaluate $\widehat{f}$ to report loss and fairness metrics $\max_{a,a'\in\mathcal{A}}\big|G_{k,a}(\widehat{f})-G_{k,a'}(\widehat{f})\big|$ for $k\in[K]$ .

\State \textbf{Return} $\{\widetilde{\rho}_a\},\widehat{f},$ and test metrics.
\end{algorithmic}
\end{algorithm}

\begin{algorithm}
\caption{ \texttt{RegionSearch}}\label{alg:region-search}
\begin{algorithmic}[1]
\Require Groups $\mathcal{A}$; hull supports $\{\widehat{\mathcal{H}}_a\}$; plug-in $\widehat{\gamma}_a,\widehat{u}_{k,a},\widehat{v}_{k,a}$;
tolerances $\{\delta_k\}$; index sets $\mathcal{K}_{\mathrm{L}},\mathcal{K}_{\mathrm{LF}}$;
centroid intervals $\{\mathcal{Q}_k\}_{k\in\mathcal{K}_{\mathrm{LF}}}$; margins $\{\varepsilon_k>0\}$.
\State $ \text{best} \gets +\infty, \vec{q}_{\mathrm{opt}} \gets \text{None},\{\widetilde\lambda_{a,j}\}\gets \text{None}, \{\widetilde \rho_a\}\gets \text{None}$.
\State Construct $\widehat{r}_a^{(j)}=(\widehat{\TPR}_a^{(j)},\widehat{\FPR}_a^{(j)},1)^\top$ from hull supports $\widehat{\mathcal{H}}_a=\{(\widehat{\TPR}_a^{(j)},\widehat{\FPR}_a^{(j)})\}_{j\in S_a}$.
\For{$\vec{q}=(q_k)_{k\in\mathcal{K}_{\mathrm{LF}}} \in \prod_{k\in\mathcal{K}_{\mathrm{LF}}}\mathcal{Q}_k$} \Comment{e.g., coarse grid}
  \State \textbf{Inner LP:} $\min_{\{\lambda_{a,j}\},\{q_\ell\}_{\ell\in\mathcal{K}_{\mathrm{L}}}}
\sum_{a\in\mathcal{A}} \left\langle \widehat{\gamma}_a,\widehat{\rho}_a \right\rangle$
  \[
  \begin{aligned}
  \text{s.t.}\quad
    &\left\{
      \begin{aligned}
        \sum_{j\in S_a}\lambda_{a,j} &= 1, \quad \lambda_{a,j}\ge 0 && \forall j\in S_a,\\
        \widehat{\rho}_a &= \sum_{j\in S_a}\lambda_{a,j}\cdot \widehat{r}_a^{(j)} &&
      \end{aligned}
     \right.
     && \forall a\in\mathcal{A}
     \\[4pt]
    & -\tfrac{\delta_\ell}{2}\le\big\langle \widehat{u}_{\ell,a},\widehat{\rho}_a\big\rangle - q_\ell\le\tfrac{\delta_\ell}{2}
     && \forall a\in\mathcal{A},\ell\in\mathcal{K}_{\mathrm{L}}
     \\[4pt]
    &\left\{
      \begin{aligned}
        \big\langle \widehat{u}_{k,a},\widehat{\rho}_a\big\rangle
        - \Big(q_k+\tfrac{\delta_k}{2}\Big)\big\langle \widehat{v}_{k,a},\widehat{\rho}_a\big\rangle &\le 0,\\
        \Big(q_k-\tfrac{\delta_k}{2}\Big)\big\langle \widehat{v}_{k,a},\widehat{\rho}_a\big\rangle
        - \big\langle \widehat{u}_{k,a},\widehat{\rho}_a\big\rangle &\le 0,
      \end{aligned}
     \right.
     && \forall a\in\mathcal{A}, k\in\mathcal{K}_{\mathrm{LF}}
     \\[4pt]
    & \left\langle \widehat{v}_{k,a},\widehat{\rho}_a\right\rangle\ge\varepsilon_k
     && \forall a\in\mathcal{A},k\in\mathcal{K}_{\mathrm{LF}}.
  \end{aligned}
  \]
  \State Let $\widehat{\Phi}(\vec{q})$ be the optimal value and $\{\lambda_{a,j}\},\{\widehat{\rho}_a\}$ the optimal variables.
  \If{$\widehat{\Phi}(\vec{q})<\text{best}$} $\text{best}\gets\widehat{\Phi}(\vec{q})$, $\vec{q}_{\mathrm{opt}}\gets\vec{q}$, $\widetilde \lambda_{a,j}\gets\lambda_{a,j}$, $ \widetilde{\rho}_a\gets\widehat{\rho}_a$.
  \EndIf
\EndFor
\State \textbf{Return} Return the target operating points $\{\widetilde{\rho}_a\}_{a\in\mathcal{A}}$.
\end{algorithmic}
\end{algorithm}

\begin{algorithm}
\caption{ \texttt{RegionSearch-FeasibilityGuard}}\label{alg:region-search-feasibility-guard}
\begin{algorithmic}[1]
\Require Groups $\mathcal{A}$; hull supports $\{\widehat{\mathcal{H}}_a\}$; plug-in $\{\widehat{\gamma}_a\},\{\widehat{u}_{k,a}\},\{\widehat{v}_{k,a}\}$;
tolerances $\vec{\delta}=\{\delta_k\}$; index sets $\mathcal{K}_{\mathrm{L}},\mathcal{K}_{\mathrm{LF}}$;
centroid intervals $\{\mathcal{Q}_k\}_{k\in\mathcal{K}_{\mathrm{LF}}}$; margins $\{\varepsilon_k>0\}$;
bisection tolerance $\tau_\alpha>0$.
\Ensure Minimal feasible expansion $\widetilde \alpha$ and operating points $\{\widetilde{\rho}_a\}_{a\in\mathcal{A}}$.

\Statex \textbf{Step 0: Instantiate search interval.}
\For{each $a\in\mathcal{A}$}
  \State $\widetilde j_a \in \arg\min_{j\in S_a}\langle \widehat{\gamma}_a,\widehat{r}_a^{(j)}\rangle$; set $\widehat{\rho}_a^{\mathrm{base}} \gets \widehat{r}_a^{(\widetilde j_a)}$.
\EndFor
\State Compute baseline metrics:
$\displaystyle g_{\ell,a}^{\mathrm{base}}=\langle \widehat{u}_{\ell,a},\widehat{\rho}_a^{\mathrm{base}}\rangle$ for $\ell\in\mathcal{K}_{\mathrm{L}}$,
$\displaystyle r_{k,a}^{\mathrm{base}}=\frac{\langle \widehat{u}_{k,a},\widehat{\rho}_a^{\mathrm{base}}\rangle}{\langle \widehat{v}_{k,a},\widehat{\rho}_a^{\mathrm{base}}\rangle}$ for $k\in\mathcal{K}_{\mathrm{LF}}$ with $\langle \widehat{v}_{k,a},\widehat{\rho}_a^{\mathrm{base}}\rangle\ge \varepsilon_k$.
\State Baseline disparities:
$\displaystyle \Delta_\ell=\max_a g_{\ell,a}^{\mathrm{base}}-\min_a g_{\ell,a}^{\mathrm{base}},\quad
\Delta_k=\max_a r_{k,a}^{\mathrm{base}}-\min_a r_{k,a}^{\mathrm{base}}.$
\State Minimal per-metric expansions:
$\displaystyle \alpha_\ell^{\min}=\Delta_\ell/\delta_\ell,\quad \alpha_k^{\min}=\Delta_k/\delta_k.$
\State Set bracket:
$\displaystyle \alpha_{\textrm{lo}}\gets 1,\quad
\alpha_{\textrm{hi}}\gets \max\Big\{1,\max_{\ell\in\mathcal{K}_{\mathrm{L}}}\alpha_\ell^{\min},\max_{k\in\mathcal{K}_{\mathrm{LF}}}\alpha_k^{\min}\Big\}.$

\Statex \textbf{Step 1: Exit early at nominal tolerances.}
\State Run \texttt{RegionSearch} with $\vec{\delta}(1)=\vec{\delta}$; 
\If{feasible}
  \State \Return $\widetilde \alpha=1$ and the returned $\{\widetilde{\rho}_a\}_{a\in\mathcal{A}}$.
\EndIf

\Statex \textbf{Step 2: Bisection-search on $\alpha$.}
\While{$\alpha_{\textrm{hi}}-\alpha_{\textrm{lo}}>\tau_\alpha$}
  \State $\alpha \gets (\alpha_{\textrm{lo}}+\alpha_{\textrm{hi}})/2$; set $\vec{\delta}(\alpha)=\{\alpha\delta_k\}$.
  \State Run \texttt{RegionSearch} with $\vec{\delta}(\alpha)$.
  \If{feasible}
    \State $\alpha_{\textrm{hi}}\gets \alpha$; cache current solution $\{\widehat{\rho}_a\}_{a\in\mathcal{A}}$.
  \Else
    \State $\alpha_{\textrm{lo}}\gets \alpha$.
  \EndIf
\EndWhile

\State \Return $\widetilde \alpha\gets \alpha_{\textrm{hi}}$ and the cached solution $\{\widetilde{\rho}_a\}_{a\in\mathcal{A}}$ at $\alpha_{\textrm{hi}}$ 
\end{algorithmic}
\end{algorithm}

\subsubsection{Interventions}
\label{sec:interv}

Using the output of Algorithm~\ref{alg:region-search-feasibility-guard}, we seek to construct classifiers that achieve target operating characteristics. 
Since there are many classifiers that can achieve any operating characteristic, the practitioner can impose additional desiderata such as minimizing the expected number of interventions (see Definition \ref{def:intervention} below).
We provide two constructions in this work, 
though we leave the full question of 
optimally designing these particular classifiers as future work, since this requires answering broader questions on a sociotechnical level.

Concretely, we provide a modular meta-algorithm that accepts any constructed classifier (Algorithm \ref{alg:construct-classifier}). 
For the additional desideratum of minimizing the expected number of interventions,
we provide two randomized classifiers applied to a mixed-GWTR: 
\texttt{AntiDiagonal} mixes the mixed-GWTR predictions with a fully random classifier on a random slice of the population, while \texttt{LabelFlipping} flips the mixed-GWTR labels with outcome-dependent probabilities.
We begin by defining interventions.


\begin{definition}[Interventions]\label{def:intervention}
Fix a baseline post-processor $f^{(0)}:\mathcal{X}\times\mathcal{A}\to\{0,1\}$ as per \eqref{eq:post-processor}.
For any candidate post-processor $f$, the population \emph{intervention rate} is
$\mathrm{Int}(f;f^{(0)}):=\Pr\left(f(X,A)\neq f^{(0)}(X,A)\right)$,
with group-wise intervention rates $\mathrm{Int}_a(f;f^{(0)}) := \Pr\left(f(X,A)\neq f^{(0)}(X,A)\mid A{=}a\right)$.
On a dataset $\mathcal{D}$, the \emph{expected empirical intervention rate}\footnote{To clarify, we are computing an expectation over the randomization procedure induced by post-processors $f$ and/or $f^{(0)}$ for a \textit{fixed} dataset $\mathcal{D}$.} is
\begin{align*}
    \widetilde{\mathrm{Int}}(f;f^{(0)};\mathcal{D}):=\frac{1}{|\mathcal{D}|}\sum_{(x,a,y)\in\mathcal{D}}\Pr\left(f(x,a)\neq f^{(0)}(x,a)\right),
\end{align*}
where the probability is taken over any randomization introduced by $f$ and/or $f^{(0)}$.
The expected number of interventions is $|\mathcal{D}|\cdot \widetilde{\mathrm{Int}}$.

Similarly, on a dataset $\mathcal{D}$, the \emph{empirical intervention rate} removes the expectation over the randomization procedures induced by the post-processors, and is given by
\begin{align*}
    \widehat{\mathrm{Int}}(f;f^{(0)};\mathcal{D}):=\frac{1}{|\mathcal{D}|}\sum_{(x,a,y)\in\mathcal{D}}1\left(f(x,a)\neq f^{(0)}(x,a)\right),
\end{align*}
so that the empirical number of interventions is $|\mathcal{D}|\cdot \widehat{\mathrm{Int}}$. 
\end{definition}
The intervention rate is simply the misclassification error when the true labels are provided by the baseline post-processor $f^{(0)}$.  However, we find it helpful to define this quantity separately.\footnote{
We do not claim that interventions are the best desiderata to consider when developing classifiers for real-world deployment. We provide a general procedure for constructing classifiers to achieve any target operating characteristic realizable by post-processing (see Algorithm~\ref{alg:construct-classifier}),
and
leave the question of designing additional ethically and legally sound objectives to practitioners (see, e.g., \citet{heidari2019long, regoli2023fair}).}\label{ftnoote:ethical-legal-optimality}

\begin{definition}[Minimal intervention]\label{def:intervention-obj}
 Consider a class $\mathcal{F}_N$ of post-processors as per \eqref{eq:post-processor}.
We define the following minimal intervention post-processing objective:
\begin{equation}\label{eq:min-int}
\min_{f\in\mathcal{F}_N}\widetilde{\mathrm{Int}}(f;f^{(0)};\mathcal{D}_{\mathrm{post}})
\quad\text{s.t.}\quad
\widehat{\rho}_a(f) =\widetilde{\rho}_a,\forall a\in\mathcal{A},
\end{equation}
where $\widehat{\rho}_a(f)$ are the empirical operating characteristic rates of post-processors $f\in\mathcal{F}_N$ on a post-processing set $\mathcal{D}_{\mathrm{post}}$ (see \S\ref{sec:region-search}),
and
$\widetilde{\rho}_a=(\widetilde{\TPR}_a,\widetilde{\FPR}_a,1)^\top$ are target operating characteristics (e.g., the output of Algorithm \ref{alg:region-search-feasibility-guard}).

\end{definition}

\subsubsection{Label Flipping}
\label{lf}

\texttt{LabelFlipping} flips the mixed-GWTR labels with outcome-dependent probabilities
$\widetilde{p}_{a,y}:=\Pr(\widetilde{f}=1\mid A=a,f^{(0)}=y)$, $y\in\{0,1\}$. The final prediction is distributed as a Bernoulli random variable with success probability 
\begin{equation}\label{eq:labelflipping-form}
    \Pr\left(\widetilde{f}(x,a)=1\mid X=x,A=a\right)=\widetilde{p}_{a,0}\left(1-q_a(x)\right)+\widetilde{p}_{a,1} q_a(x).
\end{equation}
This induces a linear map on the operating characteristics:
\begin{equation}\label{eq:labelflipping-rates}
    {\widetilde\TPR}_a =\widetilde{p}_{a,1}{\TPR}^{(0)}_a + \widetilde{p}_{a,0}\left(1-{\TPR}^{(0)}_a\right),
    ~{\widetilde\FPR}_a =\widetilde{p}_{a,1}{\FPR}^{(0)}_a + \widetilde{p}_{a,0}\left(1-{\FPR}^{(0)}_a\right).
\end{equation}

Similar to \texttt{AntiDiagonal}, fixing the baseline post-processor and the target operating points uniquely determines the free parameters of the randomization procedure (i.e., the flipping probabilities $\widetilde{p}_{a,y}$ for \texttt{LabelFlipping}).

Geometrically, the set of operating characteristics attainable by \eqref{eq:labelflipping-rates} is the triangle spanned by the baseline hull point $(\TPR^{(0)}, \FPR^{(0)})$ and the trivial classifiers $(\tpr,\fpr)=(1,1)$ and $(0,0)$ 
(equivalently $(0,1)$ and $(1,0)$ in the $(\FNR,\FPR)$-plane).

For \texttt{LabelFlipping}, when the baseline post-processor $f^{(0)}$ takes the form of \eqref{eq:baseline-postprocessor},
the linear map from \eqref{eq:labelflipping-rates} gives the $2\times2$ system
\begin{equation}\label{m}
    \begin{bmatrix}
        \widetilde{\FPR}_a \\
        1-\widetilde{\FNR}_a
    \end{bmatrix}
    =
    \begin{bmatrix}
        \FPR^{(0)}_{a,\theta} & 1-\FPR^{(0)}_{a,\theta} \\
        1-\FNR^{(0)}_{a,\theta} & \FNR^{(0)}_{a,\theta}
    \end{bmatrix}
    \begin{bmatrix}
        p_{a,1}(\theta) \\
        p_{a,0}(\theta)
    \end{bmatrix},
\end{equation}
whose determinant is $\det(\theta)=\FPR^{(0)}_{a,\theta}+\FNR^{(0)}_{a,\theta}-1$.

Similar to \texttt{AntiDiagonal}, infeasibility only occurs when the post-processor has a degenerate ROC curve.
In \eqref{m}, if $\det(\theta)\neq 0$,
\begin{align}\label{eq:labelflipping-closed-form}
p_{a,1}(\theta)&=\frac{\widetilde{\FPR}_a\FNR^{(0)}_{a,\theta} - \left(1-\widetilde{\FNR}_a\right)\left(1-\FPR^{(0)}_{a,\theta}\right)}{\det(\theta)}, \nonumber \\
p_{a,0}(\theta)&=\frac{\left(1-\widetilde{\FNR}_a\right)\FPR^{(0)}_{a,\theta} - \widetilde{\FPR}_a\left(1-\FNR^{(0)}_{a,\theta}\right)}{\det(\theta)},
\end{align}
with feasibility requiring $0\le p_{a,0}(\theta),p_{a,1}(\theta)\le 1$.

\subsubsection{Formulating the MinIntervention problem}\label{sec:minintervention-problem}

Given a fixed post-processing set, the expected number of interventions (\eqref{def:intervention}) can be computed in closed form for any mixed-GWTR. 
In particular, for a base post-processor $f^{(0)}$, let $s_a^+ := \Pr\left(f^{(0)}{=}1\mid A{=}a\right)$ denote the baseline selection rate. Then, the expected empirical intervention rates for a post-processor $\widetilde f$ on $f^{(0)}$ under our randomization procedures are given as follows.
\begin{itemize}
    \item \texttt{AntiDiagonal}: Let $(\lambda_a,p_a)$ be chosen to match the target rates via \eqref{eq:antidiagonal-rates}.
Since we flip the baseline with probability $\lambda_a$ and then draw a fresh $\mathrm{Bernoulli}(p_a)$ random variable, 
an intervention occurs exactly when we flip and the fresh draw disagrees with the baseline label. 
Thus,
\begin{equation}\label{eq:antidiagonal-intervention}
    {\mathcal{I}}_a^{\mathrm{AD}}:={\mathrm{Int}}^{\mathrm{AD}}_a(\widetilde f; f^{(0)})
    = \mathbb{E}\left[1\left(\widetilde{f}\neq f^{(0)}\right)\middle|A{=}a\right]
    = \lambda_a\left( s_a^+ (1-p_a) + (1-s_a^+) p_a\right).
\end{equation}
On a finite split $\mathcal{D}_{\mathrm{post}}^{(a)}$ with $n_a$ instances and $n^{(0)}_{1,a}$ baseline positives, this can be estimated by 
\begin{align*}
\widetilde{\mathcal{I}}_a^{\mathrm{AD}}:=\widetilde{\mathrm{Int}}^{\mathrm{AD}}_a(\widetilde f; f^{(0)}; \mathcal{D}_{\text{post}}) = \lambda_a\left(n^{(0)}_{1,a}(1-p_{a}) + (n_a-n^{(0)}_{1,a})p_{a}\right)/n_a.
\end{align*}

    \item \texttt{LabelFlipping}: 
     Inspired by the label flipping procedure\footnote{Developed for the case of perfect fairness for equality of opportunity and equalized odds.} of \citet{hardt2016equality} 
    let $(p_{a,0},p_{a,1})$ be chosen to match the target rates via \eqref{eq:labelflipping-rates}.
The expected empirical intervention rate in group $a$ is then
\begin{equation}\label{eq:labelflipping-intervention}
{\mathcal{I}}_a^{\mathrm{LF}}:={\mathrm{Int}}^{\mathrm{LF}}_a(\widetilde f; f^{(0)})
= \mathbb{E}\left[1\left(\widetilde{f}\neq f^{(0)}\right)\middle|A{=}a\right]
= s_a^+(1-p_{a,1}) + (1-s_a^+)p_{a,0}.
\end{equation}
On a finite split $\mathcal{D}_{\mathrm{post}}^{(a)}$, this can be estimated by 
\begin{align*}
    \widetilde{\mathcal{I}}_a^{\mathrm{LF}}:= \widetilde{\mathrm{Int}}^{\mathrm{LF}}_a(\widetilde f; f^{(0)}; \mathcal{D}_{\text{post}}) = \left(n^{(0)}_{1,a}(1-p_{a,1}) + (n_a-n^{(0)}_{1,a})p_{a,0}\right)/n_a.
\end{align*}
\end{itemize}

Taking the base post-processor $f^{(0)}$ to be of the form given by \eqref{eq:baseline-postprocessor}, we can
substitute the closed-form expressions for the parameters of the randomization procedure (\eqref{eq:antidiagonal-closed-form}, \eqref{eq:labelflipping-closed-form}) into the closed-form intervention expressions (\eqref{eq:antidiagonal-intervention}, \eqref{eq:labelflipping-intervention}).
This yields one-dimensional objectives in $\theta$:
\begin{align}
\mathcal{I}^{\mathrm{AD}}_{a,\theta}
&= \lambda_{a,\theta}\left(s_a^+ \left(1-p_{a,\theta}\right) + \left(1-s_a^+\right) p_{a,\theta}\right),\label{eq:intervention-antidiagonal-theta}\\
\mathcal{I}^{\mathrm{LF}}_{a,\theta}
&= s_a^+\left(1-p_{a,1}(\theta)\right) + \left(1-s_a^+\right)p_{a,0}(\theta). \label{eq:intervention-labelflipping-theta}
\end{align}

This suggests enumerating adjacent hull pairs $\mathsf{Edges}_a=\{(h,h{+}1)\}$ for each group $a\in\mathcal{A}$, and minimizing the expected empirical intervention rate 
$\widetilde{\mathcal{I}}_{a,\theta}^{\mathrm{AD/LF}}$ over $\theta\in[0,1]$ for each edge, either with a coarse grid or golden-section search (see e.g., Section 7.2 of \citet{chong2023introduction}). 
We then select the edge and $\theta$ with the smallest empirical intervention, 
thereby yielding the minimal-intervention instantiation of the chosen randomization mechanism while matching the target operating characteristics. 

Since not all operating characteristics are attainable by applying a randomization rule to a base post-processor of the form in \eqref{eq:baseline-postprocessor}, 
we check for feasibility by ensuring 
that the solved randomization parameters lie in the appropriate set (i.e, $(\lambda_a,p_a)\in[0,1]^2$ for \texttt{AntiDiagonal} and $(p_{a,0},p_{a,1})\in[0,1]^2$ for \texttt{LabelFlipping}). We skip values of $\theta$ if they are infeasible. 

\paragraph{Snapping to the convex hull boundary.} In practice, for stability and computational efficiency, we check whether the target operating characteristic for a group lies near or on its group-wise ROC convex hull before conducting the search over adjacent hull pairs.

If the target operating characteristic lies exactly on the boundary of the convex hull, we set the intervention rate to zero and skip the edge search procedure for this group.
We check closeness of the target point
within adaptive tolerance values---$\textrm{tol}_{\textrm{fpr}}{=}\xi/n_{a,0}$ and $\textrm{tol}_{\textrm{fnr}}{=}\xi/n_{a,1}$ for tunable $\xi$ where $n_{a,0}$ and $n_{a,1}$ are the number of negative and positive labels in group $a$, respectively, cf. \S\ref{sec:region-search}.
If the target point is within these tolerance levels to the convex hull,
we snap to the nearest edge and, again, skip the edge search procedure.

Altogether, this procedure is detailed in Algorithm \ref{alg:minintervention} and provides the final module of the post-processing procedure outlined in our ROCF-pipeline (Algorithm \ref{alg:rocf-pipeline}).

\subsubsection{Multiple protected attributes}\label{sec:exp:multiple-protected-attributes}

In \texttt{RegionSearch} (see Algorithm \ref{alg:region-search}), the inner LP optimizes group-wise operating characteristics over the empirical ROC-hull supports, yielding $\mathcal{O}(|\mathcal{A}|)$ linear constraints and $\mathcal{O}(\sum_{a\in\mathcal{A}} |S_a|)$ variables, where $|S_a|$ is the number of support points in the empirical convex hull of group $a$. 
As discussed in Remark \ref{remark:region-search-size-LP} and as observed in our experiments (see runtimes in \S\ref{app:add-exp:runtimes}), the set of support points is modestly sized. 
Given the optimal operating characteristics, our classifier construction decouples across groups, so the randomization hyperparameters of Algorithm \ref{alg:construct-classifier} can be found in parallel. 

\subsection{Theoretical Details and Proofs}\label{app:proofs}

\begin{definition}[Operating characteristic feasibility region]\label{def:feasibility-region}
Let $\mathcal{K}_{\mathrm{L}}$ and $\mathcal{K}_{\mathrm{LF}}$ index linear and linear–fractional group performance functions, respectively, 
and let $\vec{q}=(q_k)_{k\in\mathcal{K}_{\mathrm{LF}}}$ be fixed centroids for the LF constraints.
Define $\mathfrak{F}(\vec{q};s,\vec{\delta})$ to be the set of $\Big\{ \{\vec{\rho}_a\}_{a\in\mathcal{A}}$ such that
there are $q_\ell$, $\ell \in \mathcal{K}_{\mathrm{L}}$,
such that for all $a$
and $\ell \in \mathcal{K}_{\mathrm{L}}$,
$ -\tfrac{\delta_\ell}{2} \le 
   \langle \vec{u}_{\ell,a}, \vec{\rho}_a \rangle - q_\ell 
   \le \tfrac{\delta_\ell}{2}$.
Also, 
for all $a$
and $k \in \mathcal{K}_{\mathrm{LF}}$,
$  \langle \vec{u}_{k,a}, \vec{\rho}_a \rangle 
     - (q_k+\tfrac{\delta_k}{2}) \langle \vec{v}_{k,a}, \vec{\rho}_a \rangle \le 0$,
$ (q_k-\tfrac{\delta_k}{2}) \langle \vec{v}_{k,a}, \vec{\rho}_a \rangle     - \langle \vec{u}_{k,a}, \vec{\rho}_a \rangle \le 0$, 
$  \langle \vec{v}_{k,a}, \vec{\rho}_a \rangle > 0\Big\}$.
\end{definition}

The (centroid-free) \textit{operating characteristic feasibility region} is then the union $\mathfrak{F}(s,\vec\delta) := \bigcup_{\vec{q}\in\mathcal{Q}} \mathfrak{F}(\vec{q};s,\vec\delta)$, where $\mathcal{Q}$ is an appropriate set of candidates for the LF centroids (cf. \S\ref{sec:centroid-linearization}). 

\subsubsection{Proof of Theorem \ref{thm:centroid}}

    Denote the feasible set of the post-processing problem with fairness constraints encoded as pairwise disparities (cf. \eqref{eq:post-processing-function-opt-problem} and the discussion below it) as $\mathfrak{F}_{0}(s,\vec\delta)$.

    We prove this theorem in two parts: (i) first, we argue 
    that $\mathfrak{F}_{0}(s,\vec\delta)=\mathfrak F(s,\vec\delta)$ (cf. Definition~\ref{def:feasibility-region}); 
    then, (ii) we show that the optimal values are the same.

    \emph{(i) Equality of feasible sets.} 
    
    {\bf Part 1.} The first required inclusion is that
    $\mathfrak{F}_{0}(s,\vec\delta)\subseteq \mathfrak{F}(s,\vec\delta)$.

    Suppose $\{\vec{\rho}_a\}_{a\in\mathcal A}\in\mathfrak{F}_0(s,\vec\delta)$. 
    Clearly, for any collection of numbers $\{z_a\}_{a\in\mathcal{A}}$, $z_a\in[0,1]$ and $\delta\ge 0$,
        \begin{align}\label{eq:pairwise-centroid-equivalence}
            \max_{a,a'}|z_a-z_{a'}|\le \delta\quad\Longleftrightarrow\quad
            \exists q\in[0,1]\text{such that}\max_a |z_a-q|\le \delta/2.
        \end{align}

    Now consider the disjoint sets of constraints, $\mathcal K_L, \mathcal{K}_{LF}$, in turn:
    \begin{itemize}
        \item 
    If $k\in\mathcal K_L$, take $\{z_a\}_{a\in\mathcal{A}}=\{\langle \vec{u}_{k,a},\vec{\rho}_a\rangle\}_{a\in\mathcal{A}}$. 
    Then the pairwise disparity constraints and \eqref{eq:pairwise-centroid-equivalence} guarantee the existence of $q_k\in \mathcal{Q}_k$ s.t.
    $-\tfrac{\delta_k}{2}\le \langle \vec{u}_{k,a},\vec{\rho}_a\rangle-q_k\le \tfrac{\delta_k}{2}$ for all $a\in\mathcal{A}$.
    This centroid is admissible since it is associated with a linear constraint. 

    \item If $k\in\mathcal K_{\mathrm{LF}}$, take 
    $\{z_a\}_{a\in\mathcal{A}} = \{G_{a,k}\}_{a\in\mathcal{A}} = \{ \frac{\langle \vec{u}_{k,a},\vec{\rho}_a\rangle}{\langle \vec{v}_{k,a},\vec{\rho}_a\rangle} \}_{a\in\mathcal{A}}$, 
    which is well-defined by the denominator positivity constraint.

    Since $\max_a G_{a,k} - \min_a G_{a,k} \leq \delta_k$ for all $k\in\mathcal K_{\mathrm{LF}} \subseteq \mathcal{K}$ and $z_a\in[0,1]$, the set
    \[
        \bigcap_{a\in\mathcal A}\big[z_a-\tfrac{\delta_k}{2},z_a+\tfrac{\delta_k}{2}\big]
        = \Big[\max_a z_a-\tfrac{\delta_k}{2},\min_a z_a+\tfrac{\delta_k}{2}\Big].
    \]
     is nonempty. Moreover, it is included in 
$\mathcal \mathcal{Q}_k=[\tfrac{\delta_k}{2},1{-}\tfrac{\delta_k}{2}]$ (admissible set of centroids cf.~\eqref{eq:admissible-centroid-def}).

Pick any $q_k\in \mathcal \mathcal{Q}_k$ in this overlap.
The pairwise disparity constraints $|z_a-q_k|\le \delta_k/2$ and denominator positivity are then equivalent to 
    \[
        \langle \vec{u}_{k,a},\vec{\rho}_a\rangle - (q_k+\tfrac{\delta_k}{2})\langle \vec{v}_{k,a},\vec{\rho}_a\rangle \le 0,\quad
        (q_k-\tfrac{\delta_k}{2})\langle \vec{v}_{k,a},\vec{\rho}_a\rangle - \langle \vec{u}_{k,a},\vec{\rho}_a\rangle \le 0
    \]
for all $a\in\mathcal{A}$. 
    
Since $q_k\in\mathcal \mathcal{Q}_k$ for all $k\in\mathcal K_{\mathrm{LF}}$, we conclude $\vec q\in\mathcal Q$. \end{itemize}
It follows that $\{\vec{\rho}_a\}_{a\in\mathcal{A}}\in \mathfrak{F}(\vec q; s,\vec\delta)$ for some admissible $\vec q$.

{\bf Part 2.} The second required inclusion is that
$\mathfrak{F}_{0}(s,\vec\delta)\supseteq \mathfrak{F}(s,\vec\delta)$.

Suppose $\{\vec{\rho}_a\}_{a\in\mathcal A}\in \mathfrak{F}(\vec q; s,\vec\delta)$ for some $\vec q\in\mathcal Q$. 
Consider again the disjoint sets of constraints $\mathcal K_L, \mathcal{K}_{LF}$, in turn.
\begin{itemize}
    \item 
If $k\in\mathcal K_L$, then $|\langle \vec{u}_{k,a},\vec{\rho}_a\rangle-q_k|\le \delta_k/2$ for all $a\in\mathcal{A}$.
Hence, for any $a,a'$,
\[
    \big|\langle \vec{u}_{k,a},\vec{\rho}_a\rangle-\langle \vec{u}_{k,a'},\vec\rho_{a'}\rangle\big|
    \le \big|\langle \vec{u}_{k,a},\vec{\rho}_a\rangle-q_k\big|+\big|\langle \vec u_{k,a'},\vec \rho_{a'}\rangle-q_k\big|
    \le \delta_k
\]
so that the pairwise disparity constraints hold.

\item If $k\in\mathcal K_{\mathrm{LF}}$, then for each $a\in\mathcal{A}$,
\[
    \langle \vec{u}_{k,a},\vec{\rho}_a\rangle - (q_k+\tfrac{\delta_k}{2})\langle \vec{v}_{k,a},\vec{\rho}_a\rangle \le 0;\qquad
    (q_k-\tfrac{\delta_k}{2})\langle \vec{v}_{k,a},\vec{\rho}_a\rangle - \langle \vec{u}_{k,a},\vec{\rho}_a\rangle \le 0.
\]
By denominator positivity, 
dividing by $V_{k,a}(\vec{\rho}_a)$ gives
    $
    q_k-\tfrac{\delta_k}{2} \le G_{k,a}(\vec{\rho}_a)\le q_k+\tfrac{\delta_k}{2},
    $
so by the triangle inequality,
    $
        |G_{k,a}(\vec{\rho}_a)-G_{k,a'}(\vec\rho_{a'})|\le \delta_k
    $
for all $a,a'$.
\end{itemize}

Therefore $\{\vec{\rho}_a\}_{a\in\mathcal{A}}\in\mathfrak{F}_0(s,\vec\delta)$.

\noindent\emph{(ii) Equality of optimal values.}
Let $J(\{\vec{\rho}_a\}_{a\in \mathcal A})$ denote the linear objective in \eqref{eq:post-processing-function-opt-problem}.
From (i) we have $\mathfrak{F}_0(s,\vec\delta)=\mathfrak F(s,\vec\delta)=\bigcup_{\vec q\in\mathcal Q}\mathfrak{F}(\vec q; s,\vec\delta)$.
Hence
\[
    \min_{\{\vec{\rho}_a\}\in \mathfrak{F}_0(s,\vec\delta)} J(\{\vec{\rho}_a\})
    \;=\;
    \min_{\{\vec{\rho}_a\}\in \cup_{\vec q\in\mathcal Q}\mathfrak{F}(\vec q; s,\vec\delta)} J(\{\vec{\rho}_a\})
    \;=\;
    \min_{\vec q\in\mathcal Q}\min_{\{\vec{\rho}_a\}\in \mathfrak{F}(\vec q; s,\vec\delta)} J(\{\vec{\rho}_a\})
    \;=\;
    \min_{\vec q\in\mathcal Q} \Phi(\vec q),
\]
since the inner problem at fixed $\vec q$ is exactly \eqref{eq:linear-opt-prob} with optimal value $\Phi(\vec q)$.
Moreover, the above argument also implies that the minimizers can be recovered as stated in the theorem.
This concludes the proof.

\subsection{Practical implementation details}\label{app:imp}

We provide here additional implementation details for the experiments in \S\ref{sec:empirical}.

\paragraph{Datasets.}
\begin{itemize}
    \item \texttt{COMPAS} \citep{LarsonAngwin2016ProPublicaTechResponse} is a recidivism dataset where the goal is to predict re-offense within two years. 
    We take race as the protected attribute, restrict to the two largest groups (African-American and Caucasian, $|\mathcal{A}|{=}2$), and perform preprocessing as in \citet{cho2020fair} (e.g., removing traffic offenses and enforcing a screening–arrest window) yielding $5,\!278$ individuals in the cleaned dataset.

    We start from the ProPublica two-year cohort and retain only the columns
\texttt{[age, c\_charge\_degree, race, sex, priors\_count}, 
\texttt{days\_b\_screening\_arrest, is\_recid, c\_jail\_in, c\_jail\_out]}. 

We remove entries with inconsistent screening/arrest timing by keeping records with \texttt{days\_b\_screening\_arrest} in $[-30,30]$ days and excluding rows with \texttt{is\_recid = -1}. 
We then drop traffic offenses (\texttt{c\_charge\_degree = "O"}). 
Next, we compute \emph{length of stay} as the day difference between \texttt{c\_jail\_out} and \texttt{c\_jail\_in}, subsequently dropping the original jail timestamp columns and the screening-arrest column. We restrict the analysis to the two largest racial groups (African-American and Caucasian),
binarize the label from \texttt{is\_recid}, and define groups by \texttt{race}. 
When training the attribute-aware classifier, we remove the sensitive column from $X$ (here, \texttt{race}) and apply one-hot encoding to all remaining categorical features; the protected attribute $A$ is later concatenated as an extra input feature.

    \item \texttt{ACSIncome} \citep{ding2021retiring} comes from the 2018 one-Year American Community Survey and contains data on a total of $1,\!664,
    \!500$ individuals.
    The goal is to predict whether an individual has an income of above or below $\$50$k. 
    We consider both a binary protected attribute ($|\mathcal{A}|{=}2$) and a multi-group setting ($|\mathcal{A}|{=}5$) by either using gender (Male and Female) or grouping by race.
    We follow the preprocessing steps of \citet{xian2024unified} for precise variable definitions and bucketing procedures.

    We use Folktables’ ACS (2018, 1-Year, person) slice and apply \texttt{adult\_filter}. The feature set includes \texttt{AGEP, COW, SCHL, MAR, OCCP, POBP, RELP, WKHP, SEX, RAC1P}, with category maps supplied for interpretability. For binary classification, the target is $\mathbf{1}\{\texttt{PINCP} > \$50{,}000\}$. We set the protected attribute to either \texttt{SEX} in the binary protected attribute setting (cf. \S\ref{sec:exp:primary-result}), or bin by \texttt{RACE} for the multiple protected attribute setting. 
    
    Races are binned as follows: {\texttt{White} (\texttt{RAC1P}{=}1), \texttt{Black or African American} (2), \texttt{American Indian or Alaska Native} (merge 3,4,5), \texttt{Asian, Native Hawaiian or Other Pacific Islander} (merge 6,7), \texttt{Other} (merge 8,9)}.
    
    We convert to a data frame with dummy variables, map group labels to integer codes, and---consistent with the attribute-aware setup---drop the sensitive columns from $X$ before concatenating the group index as an additional input feature.

\end{itemize}

\paragraph{Base classifier}
The probabilistic predictor  $s$ is a small MLP with three layers and 32 hidden nodes per hidden layer, ending in a sigmoid output. 
We train with Adam (without weight decay) and binary cross-entropy, using dataset-specific epochs, learning rates, and batch sizes listed in Table~\ref{tab:train-hparams}.

\begin{table}
\caption{ Training hyperparameters used for the probabilistic predictor  $s$.}
\label{tab:train-hparams}
\begin{center}
\begin{tabular}{lcccccc}
\multicolumn{1}{c}{\bf dataset} &
\multicolumn{1}{c}{\bf epochs} &
\multicolumn{1}{c}{\bf learning rate} &
\multicolumn{1}{c}{\bf batch size} &
\multicolumn{1}{c}{\bf hidden width} &
\multicolumn{1}{c}{\bf \# layers} &
\multicolumn{1}{c}{\bf optimizer} \\
COMPAS     & 500 & $5\times 10^{-4}$ & 2048 & 32 & 3 & Adam \\
ACSIncome  & 20  & $1\times 10^{-3}$ & 128  & 32 & 3 & Adam \\
\end{tabular}
\end{center}
\end{table}

\begin{algorithm}
\caption{ Construct Classifier (MinIntervention)}\label{alg:construct-classifier}
\begin{algorithmic}[1]
\Require probabilistic predictor  $s$; groups $\mathcal{A}$; postprocessing set $\mathcal{D}_{\mathrm{post}}$; target rates $\{\widetilde\rho_a\}_{a\in\mathcal{A}}$ from Algorithm~\ref{alg:region-search-feasibility-guard}; a construction subroutine $\mathsf{Construct}\in\{\texttt{AntiDiagonal},\texttt{LabelFlipping},\text{any admissible}\}$.
\For{each $a\in\mathcal{A}$}
  \State Extract group subset $\mathcal{D}_{\mathrm{post}}^{(a)}=\{(x_i,a,y_i)\in\mathcal{D}_{\mathrm{post}}:A_i=a\}$.
  \State \textbf{Call subroutine:} $ \{\vec\zeta_a\}_{a\in\mathcal{A}} \leftarrow \mathsf{Construct}\left(s(\cdot,a),\mathcal{D}_{\mathrm{post}}^{(a)},\widetilde{\rho}_a\right)$.
  \State Define $ f_a(\cdot)$ and $f_a^{(0)}(\cdot)$ from parameters $\vec\zeta_a$ (e.g., mixing thresholds; label-flip rates).
\EndFor
\State Assemble $f(x,a):=f_a(x)$, $f^{(0)}(x,a):=f_a^{(0)}(x)$ and compute $\widehat{\rho}_a(f)$ on $\mathcal{D}_{\mathrm{post}}$.
\State Report $\widehat{\mathrm{Int}}(f;f^{(0)};\mathcal{D}_{\mathrm{post}})$ and $\{\widehat{\mathrm{Int}}_a(f;f^{(0)};\mathcal{D}_{\mathrm{post}})\}$.
\State \textbf{Return} $f$, realized rates $\{\widehat{\rho}_a\}$, and intervention statistics.
\end{algorithmic}
\end{algorithm}

\begin{algorithm}
\caption{ \texttt{MinIntervention} subroutine (unified \texttt{AntiDiagonal}/\texttt{LabelFlipping})}\label{alg:minintervention}
\begin{algorithmic}[1]
\Require probabilistic predictor  $s$; post-processing data $\mathcal{D}_{\mathrm{post}}$; target operating characteristics $\{(\widetilde{\TPR}_a,\widetilde{\FPR}_a)\}_{a\in\mathcal{A}}$; rate tolerance $\xi$; 1D line-search routine $G$; \textsf{mode} $\in\{\texttt{AD},\texttt{LF}\}$.
\Ensure parameter recipe and mechanism-specific parameters $\{\vec\zeta_a\}_{a\in\mathcal{A}}$ 
\For{group $a\in\mathcal{A}$}

\State \textbf{Compute empirical hull.} Extract the subpopulation $\{(x_i,a,y_i)\in\mathcal{D}_{\mathrm{post}}^{(a)}\}$, and use $s(x_i,a)$ to form the empirical ROC and upper convex hull (cf.\S\ref{sec:region-search}) and order it:
\[
\widehat{\mathcal{H}}^{\textrm{ord}}_a=\left\{\left(t_{1,a},\ldots,t_{S_a,a}\right)\text{~along with~}(\widehat{\FNR}_a^{(h)},\widehat{\FPR}_a^{(h)},\widehat{s}_{a,+}^{(h)})\right\}_{h=1}^{S_a};
\]
here, $\widehat{s}_{a,+}$ is the plug-in estimate for $s_{a,+}^{(h)}:=\Pr(f^{(0)}{=}1\mid A{=}a)$ where $f^{(0)}(x,a) = 1\left(s(x,a)\ge t_{h,a}\right)$.
\State $ \text{best} \gets +\infty$.

\State \textbf{Snap to convex hull boundary.} If $(\widetilde{\TPR}_a,\widetilde{\FPR}_a)$ lies on boundary of $\widehat{\mathcal{H}}^{\textrm{ord}}_a$ with snap tolerance $\xi$ (cf. \S\ref{sec:minintervention-problem}), compute the interpolant $\theta^{\mathrm{edge}}\in[0,1]$ and set:
  \begin{itemize}
    \item \textsf{mode}=\texttt{AD}: use $\lambda_a{=}0$ (no mixing with a random flip); objective $=0$.
    \item \textsf{mode}=\texttt{LF}: use $p_{a,0}{=}0,p_{a,1}{=}1$ (no flips); objective $=0$.
  \end{itemize}
  Update $\vec\zeta_a\leftarrow\big(t_{h,a},t_{h+1,a},\theta^{\mathrm{edge}},\text{mechanism params}\big)$, $\text{best}\leftarrow 0$, and \textbf{continue} to the next group.

\For{each adjacent pair $(h,h{+}1)$ in $\widehat{\mathcal{H}}^{\textrm{ord}}_a$},  \label{line:loop-edges}
  

  
  \State \textbf{Search over $\theta$ on this edge} via a 1D line search $G$ (e.g., grid search, golden-section search):
  \begin{enumerate}
    \item Read off \[
      \widehat{\FPR}^{(0)}_{a,\theta}=(1-\theta)\widehat{\FPR}_a^{(h)}+\theta \widehat{\FPR}_a^{(h+1)},\quad
      \widehat{\FNR}^{(0)}_{a,\theta}=(1-\theta)\widehat{\FNR}_a^{(h)}+\theta \widehat{\FNR}_a^{(h+1)},
  \]
  and the baseline selection rate $\widehat{s}_{a,+}(\theta)=(1-\theta)\widehat{s}_{a,+}^{(h)}+\theta \widehat{s}_{a,+}^{(h+1)}$.
    \item \textbf{Compute mechanism parameters.}
      \begin{itemize}
        \item If \textsf{mode}=\texttt{AD}: compute $\left(\lambda_{a,\theta},p_{a,\theta}\right)$ by the closed forms in \eqref{eq:antidiagonal-closed-form}; feasibility requires  $\left(\lambda_{a,\theta},p_{a,\theta}\right) \in [0,1]^2$ 
        \item If \textsf{mode}=\texttt{LF}: compute $\left(p_{a,0}(\theta),p_{a,1}(\theta)\right)$ via \eqref{eq:labelflipping-closed-form}; feasibility requires  $\left(p_{a,0}(\theta),p_{a,1}(\theta)\right) \in [0,1]^2$.
      \end{itemize}
      Skip if infeasible.
    \item \textbf{Compute expected number of interventions.}
      \begin{itemize}
        \item \textsf{mode}=\texttt{AD}: evaluate $\widetilde{\mathcal{I}}^{\mathrm{AD}}_{a,\theta}$ (cf. \eqref{eq:intervention-antidiagonal-theta}).
        \item \textsf{mode}=\texttt{LF}: evaluate $\widetilde{\mathcal{I}}^{\mathrm{LF}}_{a,\theta}$ (cf. \eqref{eq:intervention-labelflipping-theta}).
      \end{itemize}
    \item If $\widetilde{\mathcal{I}}^{\mathrm{AD/LF}}_{a,\theta}$ $<\text{best}$, set $\vec\zeta_a\leftarrow\left( t_{h,a},t_{h+1,a},\theta,\text{mechanism params at }\theta\right)$ and $\text{best}\leftarrow \widetilde{\mathcal{I}}^{\mathrm{AD/LF}}_{a,\theta}$.
  \end{enumerate}
\EndFor
\State \textbf{Assemble outputs.} 
Return the parameter recipe
\[
\vec\zeta_a=\left( t_{h,a},t_{h+1,a},\theta,\underbrace{(\lambda_a,p_a)}_{\textsf{AD}}\text{~or~}\underbrace{(p_{a,0},p_{a,1})}_{\textsf{LF}}\right),
\]
which specifies a base post-processor $f_a^{(0)}(\cdot)$ by \eqref{eq:baseline-postprocessor} and the final $f_a(\cdot)$ via \eqref{eq:antidiagonal-form} (\textsf{AD}) or \eqref{eq:labelflipping-form} (\textsf{LF}).
\State \Return $\{\vec \zeta_a\}_{a\in\mathcal{A}}$ 
\EndFor
\end{algorithmic}
\end{algorithm}

\paragraph{Baselines.}
We compare with
two post-processing methods that seek to simultaneously control LF fairness constraints (\texttt{META} and \texttt{MFOpt}), 
and one state-of-the-art post-processing method that controls for linear fairness constraints (\texttt{LPP}). In addition, we record the performance of the unconstrained probabilistic classifier $s$ (\texttt{Baseline}) and an oracle post-processor returned by the \texttt{RegionSearch-FG} routine of Algorithm \ref{alg:region-search-feasibility-guard} (\texttt{Oracle}).
\begin{itemize}
    \item \texttt{Baseline} refers to the probabilistic predictor  $s$ with no fairness constraints imposed. 
    It does not use the POST set at all (neither for post-processing/additional calibration nor as additional training data for the TRAIN set) and instead directly evaluates on the TEST set. 

    \item \texttt{Oracle} denotes the optimal operating point returned by \texttt{RegionSearch-FG} (Algorithm \ref{alg:region-search-feasibility-guard}) using the TEST set. 
    These rates
    maximize accuracy over operating characteristics but do not correspond to a valid classifier, since they use the test data.\footnote{The optimality of this operating point depends on how refined the grid-search over the LF fairness constraints is---however, this approximation is a fundamental feature of the centroid linearization technique, see, e.g., Theorem 4.4 of \citet{celis2019classification}. Our constructed classifiers often achieve these optimal operating characteristics (\S\ref{sec:exp:primary-result}-\S\ref{sec:exp:multiple-LF-constraints}).}

    \item \texttt{META} \citep{celis2019classification} learns a deterministic GWTR (see \eqref{eq:GWTR}) that maximizes classification accuracy subject to fairness constraints involving ratios of LF/F group performance functions (cf. \eqref{def:LF-group-performance}).
    This meta-algorithm reduces these ratio-based constraints to bounds on group performance functions controlled by a hyperparameter $\tau$; 
    we tune $\tau$ on POST and evaluate the fixed rule on TEST.

    \item \texttt{MFOpt} \citep{hsu2022pushing} learns a group-conditional randomized post-processor—i.e., a label-flipping rule parameterized by a $2{\times}2$ transition matrix per group that maps base predictions to final labels as in \citet{hardt2016equality}. 
    This method seeks to minimize the expected number of flipped labels (see also \S\ref{sec:construct-classifiers}) 
    subject to the fairness constraints on POST. The learned mapping is then fixed and applied to TEST.

    \item \texttt{LPP-DP/EOpp/EO} \citep{xian2024unified} learns a linear post-processor of the base probabilistic predictor  $s$ that satisfies $\vec\delta$-approximate fairness for common linear metrics (see \S\ref{sec:setup}).
    Though it can handle multiple classes $|\mathcal{Y}|>2$, specializing it to the binary setting results in a deterministic GWTR, similar to \citet{celis2019classification}.
    \texttt{LPP-DP/EO} refers to either imposing $\delta$-approximate demographic parity, equality of opportunity, or equalized odds, respectively.
\end{itemize}

\paragraph{Baseline configurations}
\begin{itemize}
    \item \texttt{META} \citep{celis2019classification}: We implement the group-fair reduction with simultaneous DP, EO (TPR/FPR), and PP. 
    A \emph{ratio band} of width $\tau$ is enforced by sweeping the lower endpoint $a$ on a grid with step $\varepsilon=0.01$ and setting the upper endpoint to $\min(1,a/\tau)$. 
    We select the band by first determining whether the resulting fairness constraints are satisfied on the POST set, and then choosing the band with the highest accuracy. 
    Finally, we deploy the resulting deterministic score rule on TEST. 
    We use $\tau\in\{0.1,0.2,\ldots,1.0\}$ unless stated otherwise.

    \item \texttt{MFOpt} \citep{hsu2022pushing}: We export the probabilistic predictor 's scores on POST/TEST, 
    run their provided solver on POST to obtain a stochastic matrix mapping the transition rates between bins. 
    Since the optimizer sometimes does not return a row-statistic matrix mapping, we  
    normalize it to ensure a valid classifier. 
    We use this normalized mapping on TEST and report the sampled metrics from the randomized transitions.

    \item \texttt{LPP-DP/EOpp/EO} \citep{xian2024unified}: We use the authors’ Linear Post-Processing algorithm (\texttt{LPP}) to solve the empirical LP on POST with tolerance $\alpha{=}\delta$ under either demographic parity (\texttt{DP}), equality of opportunity (\texttt{EOpp}) or equalized odds (\texttt{EO}).
    The resulting (deterministic) decision rule is obtained by linearly adjusting per-class risks, and is then applied unchanged to TEST to report accuracy and disparities.
    We use CVXPY with GUROBI (fallback SCS) and do not sweep any hyperparameters beyond the nominal $\delta$. 
\end{itemize}

{\bf Sources of randomness.}
The stochasticity of the entire procedure is induced by the random TRAIN/POST/TEST split, 
the stochastic training of $s$, 
and any randomness introduced by sampled predictions (e.g., \texttt{MFOpt} of \citet{hsu2022pushing} and our methods \texttt{AntiDiagonal} and \texttt{LabelFlipping}). 

\paragraph{Configurations for \texttt{ROCF-AD/LF} (our methods).}
For the LF-fairness constraint grid $\mathcal{Q}_{\textrm{PP}}$ of the \texttt{RegionSearch} subroutine (Algorithm~\ref{alg:region-search}), 
we sweep across $q{=}1000$ equidistantly spaced points within the range of the admissible centroids.
For the setting with multiple LF-fairness constraints, we sweep across $q{=}100$ equidistant points within the admissible bands $\mathcal{Q}_{\textrm{PP}}$ and $\mathcal{Q}_{\textrm{FOR}}$ instead.
The denominator positivity margins $\varepsilon_k$ are uniformly set to $10^{-7}$; see \S\ref{app:imp:centroids} for full details.

We set the bisection search tolerance of the wrapper algorithm \texttt{RegionSearch-FG} (Algorithm~\ref{alg:region-search-feasibility-guard}) to $\tau_{\alpha}{=}0.01$.

When constructing the minimum intervention classifier (Algorithm~\ref{alg:minintervention}), we set the snap tolerance parameter $\xi$ to be $0.75$ for both \textsc{AntiDiagonal} and \textsc{LabelFlipping}.
For both modalities, we use a golden-section search by first evaluating a coarse 101-point uniform grid over the mixing parameter $\theta$ and then, for contiguous feasible intervals, running a golden-section search
with tolerance $10^{-5}$ and a cap of 40 iterations.

On a computational level, incorporating an additional LF constraint
amounts to an extra (low-dimensional) grid search in \texttt{RegionSearch} (Algorithm \ref{alg:region-search}).
Though this yields a polynomial-time search in the grid size, we find that a modest, refined grid of $q{=}100$ equidistant points for both FOR-parity and PP performs well in practice (see \S\ref{app:imp:centroids} and \S\ref{app:add-exp:runtimes}).

\subsubsection{Admissible centroids for LF constraints}\label{app:imp:centroids}

\paragraph{Admissible centroid sets.}
For each linear–fractional (LF) constraint $k\in\mathcal K_{\mathrm{LF}}$, let
\(
[L_k,U_k]\subseteq[0,1]
\quad\text{for all }a\in\mathcal A,
\)
denote the range of the LF group performance function (cf. Definition~\ref{def:LF-group-performance}). 
Given a disparity level $\delta_k\in[0,U_k-L_k]$, we define the \emph{admissible centroid interval} as
\begin{equation}\label{eq:admissible-centroid-def}
    \mathcal \mathcal{Q}_k \;:=\; \left[L_k+\tfrac{\delta_k}{2},U_k-\tfrac{\delta_k}{2}\right].
\end{equation}

Denote $z_a:=G_{k,a}(\vec{\rho}_a)$, $a\in\mathcal A$,
which satisfy 
$\max_{a,a'}|z_a-z_{a'}|\le \delta_k$.
Then denote
\[
    \bigcap_{a\in\mathcal A}\big[z_a-\tfrac{\delta_k}{2}, z_a+\tfrac{\delta_k}{2}\big]
    =\left[\max_a z_a -\tfrac{\delta_k}{2}, \min_a z_a+\tfrac{\delta_k}{2}\right] =:\mathcal{I}_k,
\]
so that
\(
    \mathcal{I}_k\cap
    \left[L_k+\tfrac{\delta_k}{2},U_k-\tfrac{\delta_k}{2}\right]
    \neq \varnothing,
\)
and any centroid $q_k$ chosen from the intersection automatically lies in $\mathcal \mathcal{Q}_k$.

For common LF constraints like predictive parity and false omission rate parity, 
we clearly have 
$[L_k,U_k]=[0,1]$ and $\delta_k\in[0,1]$.
Hence, the admissible set of centroids for these metrics is
$\mathcal \mathcal{Q}_k=[\delta_k/2,1{-}\delta_k/2]$.

\paragraph{\bf Theoretical bands for common LF constraints.}
Let $\pi_a{=}\Pr(Y{=}1{\mid}A{=}a)$ be the prevalence/base rate. The following conditions ensure \eqref{eq:lin-frac-centroid} and denominator positivity ($V_{k,a}(\rho)= \langle \vec{v}_{k,a},\rho\rangle \ge \varepsilon_k > 0$ for $k\in\mathcal{K}_{\textrm{LF}}$; see \S\ref{sec:centroid-linearization})
are met for predictive parity and false omission rate parity. 

\emph{Predictive parity (PP).}
For $G_{\mathrm{PP},a}=\tfrac{\pi_a\TPR_a}{\pi_a\TPR_a+(1-\pi_a)\FPR_a}\in(0,1)$, the centroid band
$\left|G_{\mathrm{PP},a}-q_{\mathrm{PP}}\right|\le \delta_{\mathrm{PP}}/2$
is equivalent to the linear inequalities in $(\FPR_a,\FNR_a)$:
\[
    1-\FNR_a \in\
    \left[\alpha^{\mathrm{lo}}_a(q_{\mathrm{PP}})\FPR_a, \alpha^{\mathrm{hi}}_a(q_{\mathrm{PP}})\FPR_a\right],
    \quad
    \alpha^{\mathrm{lo/hi}}_a(q)=
    \frac{(q\mp \tfrac{\delta_{\mathrm{PP}}}{2})(1-\pi_a)}{\big(1-(q\mp \tfrac{\delta_{\mathrm{PP}}}{2})\big)\pi_a},
\]
which is valid when $(1-(q\mp \delta_{\mathrm{PP}}/2))\pi_a>0$ for all $a\in\mathcal{A}$.

Denominator positivity requires $\pi_a(1-\FNR_a)+(1-\pi_a)\FPR_a \ge \varepsilon_{\mathrm{PP}} > 0$ for all $a\in\mathcal{A}$.

\emph{False omission rate (FOR).}
For $G_{\mathrm{FOR},a}=\tfrac{\pi_a\FNR_a}{(1-\pi_a)(1-\FPR_a)+\pi_a\FNR_a}\in(0,1)$,
let $L=q_{\mathrm{FOR}}-\delta_{\mathrm{FOR}}/2$ and $U=q_{\mathrm{FOR}}+\delta_{\mathrm{FOR}}/2$.
The centroid band $\left|G_{\mathrm{FOR},a}-q_{\mathrm{FOR}}\right|\le \delta_{\mathrm{FOR}}/2$
is equivalent to the two linear constraints:
\[
    (1{-}L)\pi_a\FNR_a + L(1{-}\pi_a)\FPR_a \ge L(1{-}\pi_a),\qquad
    (1{-}U)\pi_a\FNR_a + U(1{-}\pi_a)\FPR_a \le U(1{-}\pi_a).
\]

Denominator positivity requires $(1-\pi_a)(1-\FPR_a) + \pi_a\FNR_a \ge\varepsilon_{\mathrm{FOR}} > 0$ for all $a\in\mathcal{A}$.

{\bf Practical implementation of bands for common LF constraints.}
As derived above, admissible centroids require (i) the band coefficients to be well-defined 
$    \textrm{PP}:(1-(q\pm\delta_{\textrm{PP}}/2))\pi_a>0
$
and (ii) positivity of the LF denominators 
\begin{align*}
    \textrm{PP}: \pi_a(1-\FNR_a)+(1-\pi_a)\FPR_a\ge \varepsilon_{\mathrm{PP}};
    \qquad 
    \textrm{FOR}: (1-\pi_a)(1-\FPR_a)+\pi_a\FNR_a\ge \varepsilon_{\mathrm{FOR}}.
\end{align*}

In practice, we perform an outer grid search over $\mathcal{Q}_k$ and impose additional box constraints on the inner-LP of \eqref{eq:linear-opt-prob}.

\emph{Intervals and grids.}
We ensure band coefficients are well defined by restricting centroids to the outer grids,
\[
    \mathcal{Q}_{\mathrm{PP}}=[\delta_{\mathrm{PP}}/2, 1{-}\delta_{\mathrm{PP}}/2],
    \quad
    \mathcal{Q}_{\mathrm{FOR}}=[\delta_{\mathrm{FOR}}/2,
    1{-}\delta_{\mathrm{FOR}}/2].
\]

\begin{itemize}
\item If only one LF metric is active (PP or FOR), we sample a uniform grid of $1000$ points over its interval.
\item If both PP and FOR are active, we sample 100 PP points and 100 FOR points and iterate over the pairs in their Cartesian product ($100\times 100$ pairs in total, barring denominator guards).
\end{itemize}

\emph{LP box-constraints.}
To enforce denominator positivity, we add the box-constraints $\FPR_a\ge \varepsilon$, $0\le \FNR_a\le 1$ in the inner-LP of \eqref{eq:linear-opt-prob}. 
This is a slight relaxation; in practice, with $\pi_a\in(0,1)$ and our specified grids $\mathcal{Q}_{\textrm{PP}}$, $\mathcal{Q}_{\textrm{FOR}}$, these boxes are sufficient to ensure the more formal condition ($V_{k,a}(\rho)>0$) and work well in practice. 

\subsection{Additional Experimental Results}\label{app:add-exp}

\subsubsection{Additional Results for Experiments in Main Body}\label{app:add-exp:results}

We present here experimental results for the binary protected attribute setting on \textsc{ACSIncome} and the multiple linear-fractional constraint setting for both \textsc{COMPAS} and \textsc{ACSIncome} for completeness; see Tables \ref{tab:compas-acsincome-2} and \ref{tab:compas-acsincome-multiple-LF-constraints-2}.

Our methods \texttt{ROCF-AD/LF} perform favorably on \textsc{ACSIncome} with binary protected attributes for the setting where DP, EOpp, PEq, and PP are controlled for at level $0.05$ (Table~\ref{tab:compas-acsincome-2}).

For the multiple linear fractional setting, we observe that our method can recover a mixed GWTR with no added randomization as the final classifier
and 
perform as well as state-of-the-art post-processing methods (Table~\ref{tab:compas-acsincome-multiple-LF-constraints-2}\.(B)).

We also report the behavior of the feasibility guards triggered during our calls to the region search algorithm (Algorithm~\ref{alg:region-search-feasibility-guard}).
This algorithm was called on both the post-processing dataset
and on the test set (Table~\ref{tab:feasibility-guard-summary}\.(A) and (B), respectively), 
since \texttt{ROCF-AD/LF} (or \texttt{Oracle}) finds and uses optimal operating characteristics from the post-processing (test) dataset (cf. \S\ref{sec:empirical}).

The guard was rarely triggered 
across our experiments.
For the experiment which had a substantial number of triggers (\textsc{ACSIncome} with multiple protected attributes and multiple LF fairness constraints), 
we found that the resulting expansion was quite small ($\alpha{\approx}1.04$) and yielded strong empirical results (Tables \ref{tab:compas-acsincome-multiple-LF-constraints} and \ref{tab:feasibility-guard-summary}).

On a computational level, any runs that trigger the feasibility guard also require an additional bisection search on the expansion parameter $\alpha$ and thus requires multiple calls to Algorithm \ref{alg:region-search}, 
but we find that a modest level of tolerance for terminating the search ($\tau_\alpha{=}0.01$) performs well (see Tables~\ref{tab:compas-acsincome-multiple-LF-constraints} and~\ref{tab:feasibility-guard-summary}); 
for further discussion of computational runtimes, see \S\ref{app:add-exp:runtimes}.

\begin{table}
\footnotesize
\caption{ Performance on the test set for ACSIncome ($|\mathcal{A}|{=}2$). 
The disparities $\delta_{\textrm{DP}}, \delta_{\textrm{EOpp}}, \delta_{\textrm{PEq}}, \delta_{\textrm{PP}}$ are controlled at level $0.05$ whenever they are active. 
\textbf{Interv.} is the empirical intervention rate on the test set (see Definition~\ref{def:intervention}). 
Cells in green indicate that the fairness constraint is satisfied within two standard deviations at level $0.05$, whereas cells in red indicate violation. 
Entries in the  Oracle rows are shaded in lighter colors to denote that they are not practically feasible baselines. }
\label{tab:compas-acsincome-2} 
\begin{center} 

\begin{tabular}{lccccccc}
\toprule
\multicolumn{8}{c}{\bf ACSIncome ($|\mathcal{A}|{=}2$)} \\
\addlinespace[0.35em]
\multicolumn{1}{c}{\bf Method} & \multicolumn{1}{c}{\bf Acc} & \multicolumn{1}{c}{\bf DP} & \multicolumn{1}{c}{\bf EOpp} & \multicolumn{1}{c}{\bf PEq} & \multicolumn{1}{c}{\bf PP} & \multicolumn{1}{c}{\bf FOR} & \multicolumn{1}{c}{\bf Interv.} \\
\midrule
Baseline & 0.79 $\pm$ 0.01 & \rcell{0.17 $\pm$ 0.02} & \rcell{0.14 $\pm$ 0.02} & \gcell{0.08 $\pm$ 0.03} & \gcell{0.05 $\pm$ 0.01} & {0.06 $\pm$ 0.02} & 0.00 $\pm$ 0.00 \\
\addlinespace[0.25em]\cdashline{1-8}[0.4pt/1pt]\addlinespace[0.15em]
Oracle & 0.75 $\pm$ 0.00 & \gcelloracle{0.05 $\pm$ 0.00} & \gcelloracle{0.05 $\pm$ 0.00} & \gcelloracle{0.02 $\pm$ 0.00} & \gcelloracle{0.05 $\pm$ 0.00} & {0.14 $\pm$ 0.00} & N/A \\
\textbf{ROCF-AD} (ours) & 0.75 $\pm$ 0.00 & \gcell{0.05 $\pm$ 0.00} & \gcell{0.05 $\pm$ 0.00} & \gcell{0.02 $\pm$ 0.00} & \gcell{0.05 $\pm$ 0.00} & {0.14 $\pm$ 0.00} & 0.02 $\pm$ 0.00 \\
\textbf{ROCF-LF} (ours) & 0.75 $\pm$ 0.00 & \gcell{0.05 $\pm$ 0.00} & \gcell{0.05 $\pm$ 0.00} & \gcell{0.02 $\pm$ 0.00} & \gcell{0.05 $\pm$ 0.00} & {0.14 $\pm$ 0.00} & 0.02 $\pm$ 0.00 \\
\addlinespace[0.25em]\cdashline{1-8}[0.4pt/1pt]\addlinespace[0.15em]
MFOpt & 0.74 $\pm$ 0.01 & \rcell{0.20 $\pm$ 0.01} & \rcell{0.09 $\pm$ 0.01} & \rcell{0.15 $\pm$ 0.01} & \rcell{0.06 $\pm$ 0.01} & {0.07 $\pm$ 0.01} & 0.06 $\pm$ 0.01 \\
META & 0.79 ± 0.00 & \rcell{0.11 ± 0.02} & \gcell{0.03 ± 0.02} & \gcell{0.03 ± 0.01} & \rcell{0.11 ± 0.01} & 0.09 ± 0.01 & 0.00 ± 0.00 \\
\addlinespace[0.25em]\cdashline{1-8}[0.4pt/1pt]\addlinespace[0.15em]
LPP-DP & 0.78 $\pm$ 0.01 & \gcell{0.05 $\pm$ 0.00} & \gcell{0.04 $\pm$ 0.01} & \gcell{0.02 $\pm$ 0.01} & \rcell{0.14 $\pm$ 0.01} & {0.12 $\pm$ 0.01} & 0.00 $\pm$ 0.00 \\
LPP-EO & 0.78 $\pm$ 0.01 & \rcell{0.10 $\pm$ 0.01} & \gcell{0.03 $\pm$ 0.01} & \gcell{0.03 $\pm$ 0.01} & \rcell{0.10 $\pm$ 0.01} & {0.09 $\pm$ 0.01} & 0.00 $\pm$ 0.00 \\
\bottomrule
\end{tabular}

\end{center}
\end{table}

\begin{table}
\footnotesize
\caption{ Performance on the test set for (A) COMPAS ($|\mathcal{A}|{=}2$) and (B) ACSIncome ($|\mathcal{A}|{=}2$). 
The disparities $\delta_{\textrm{EOpp}}, \delta_{\textrm{PP}}$ and $\delta_{\textrm{FOR}}$ are controlled at level $0.10$ whenever they are active. 
\textbf{Interv.} is the empirical intervention rate on the test set (see Definition~\ref{def:intervention}). 
Cells in green indicate that the fairness constraint is satisfied within two standard deviations at level $0.10$, whereas cells in red indicate violation. 
Entries in the  Oracle rows are shaded in lighter colors to denote that they are not practically feasible baselines. }
\label{tab:compas-acsincome-multiple-LF-constraints-2}
\begin{center} 
\begin{tabular}{lccccccc}
\toprule
\multicolumn{8}{c}{\bf (A) COMPAS} \\
\addlinespace[0.35em]
\multicolumn{1}{c}{\bf Method} & \multicolumn{1}{c}{\bf Acc} & \multicolumn{1}{c}{\bf DP} & \multicolumn{1}{c}{\bf EOpp} & \multicolumn{1}{c}{\bf PEq} & \multicolumn{1}{c}{\bf PP} & \multicolumn{1}{c}{\bf FOR} & \multicolumn{1}{c}{\bf Interv.} \\
\midrule
Baseline & 0.68 $\pm$ 0.01 & 0.28 $\pm$ 0.05 & \rcell{0.27 $\pm$ 0.05} & 0.19 $\pm$ 0.05 & \gcell{0.07 $\pm$ 0.04} & \gcell{0.03 $\pm$ 0.02} & 0.00 $\pm$ 0.00 \\
\addlinespace[0.25em]\cdashline{1-8}[0.4pt/1pt]\addlinespace[0.15em]
Oracle & 0.65 $\pm$ 0.04 & 0.15 $\pm$ 0.04 & \gcelloracle{0.10 $\pm$ 0.01} & 0.12 $\pm$ 0.09 & \gcelloracle{0.10 $\pm$ 0.01} & \gcelloracle{0.09 $\pm$ 0.01} & N/A \\
\textbf{ROCF-AD} (ours) & 0.65 $\pm$ 0.04 & 0.15 $\pm$ 0.04 & \gcell{0.12 $\pm$ 0.03} & 0.11 $\pm$ 0.09 & \gcell{0.11 $\pm$ 0.04} & \gcell{0.08 $\pm$ 0.04} & 0.01 $\pm$ 0.02 \\
\textbf{ROCF-LF} (ours) & 0.65 $\pm$ 0.04 & 0.15 $\pm$ 0.04 & \gcell{0.11 $\pm$ 0.04} & 0.11 $\pm$ 0.09 & \gcell{0.11 $\pm$ 0.04} & \gcell{0.09 $\pm$ 0.04} & 0.01 $\pm$ 0.02 \\
\addlinespace[0.25em]\cdashline{1-8}[0.4pt/1pt]\addlinespace[0.15em]
LPP-EOpp & 0.68 $\pm$ 0.01 & 0.15 $\pm$ 0.04 & \gcell{0.14 $\pm$ 0.04} & 0.08 $\pm$ 0.04 & \gcell{0.12 $\pm$ 0.03} & \gcell{0.06 $\pm$ 0.03} & 0.00 $\pm$ 0.00 \\
\bottomrule
\end{tabular}


\begin{tabular}{lccccccc}
\toprule
\multicolumn{8}{c}{\bf (B) ACSIncome ($|\mathcal{A}|{=}2$)} \\
\addlinespace[0.35em]
\multicolumn{1}{c}{\bf Method} & \multicolumn{1}{c}{\bf Acc} & \multicolumn{1}{c}{\bf DP} & \multicolumn{1}{c}{\bf EOpp} & \multicolumn{1}{c}{\bf PEq} & \multicolumn{1}{c}{\bf PP} & \multicolumn{1}{c}{\bf FOR} & \multicolumn{1}{c}{\bf Interv.} \\
\midrule
Baseline & 0.79 $\pm$ 0.01 & 0.17 $\pm$ 0.02 & \gcell{0.14 $\pm$ 0.02} & 0.08 $\pm$ 0.03 & \gcell{0.05 $\pm$ 0.01} & \gcell{0.06 $\pm$ 0.02} & 0.00 $\pm$ 0.00 \\
\addlinespace[0.25em]\cdashline{1-8}[0.4pt/1pt]\addlinespace[0.15em]
Oracle & 0.79 $\pm$ 0.00 & 0.16 $\pm$ 0.00 & \gcelloracle{0.10 $\pm$ 0.00} & 0.07 $\pm$ 0.01 & \gcelloracle{0.07 $\pm$ 0.01} & \gcelloracle{0.07 $\pm$ 0.00} & N/A \\
\textbf{ROCF-AD} (ours) & 0.79 $\pm$ 0.00 & 0.16 $\pm$ 0.00 & \gcell{0.10 $\pm$ 0.00} & 0.07 $\pm$ 0.01 & \gcell{0.07 $\pm$ 0.01} & \gcell{0.07 $\pm$ 0.00} & 0.00 $\pm$ 0.00 \\
\textbf{ROCF-LF} (ours) & 0.79 $\pm$ 0.00 & 0.16 $\pm$ 0.00 & \gcell{0.10 $\pm$ 0.00} & 0.07 $\pm$ 0.01 & \gcell{0.07 $\pm$ 0.01} & \gcell{0.07 $\pm$ 0.00} & 0.00 $\pm$ 0.00 \\
\addlinespace[0.25em]\cdashline{1-8}[0.4pt/1pt]\addlinespace[0.15em]
LPP-EOpp & 0.79 $\pm$ 0.01 & 0.14 $\pm$ 0.01 & \gcell{0.09 $\pm$ 0.01} & 0.05 $\pm$ 0.02 & \gcell{0.08 $\pm$ 0.01} & \gcell{0.07 $\pm$ 0.01} & 0.00 $\pm$ 0.00 \\
\bottomrule
\end{tabular}
\end{center}
\end{table}

\begin{table}
\caption{ Summary statistics for feasibility guard triggers of \texttt{RegionSearch-FG} (Algorithm~\ref{alg:region-search-feasibility-guard}) for the nominal disparity levels considered in \S\ref{sec:exp:primary-result}-\S\ref{sec:exp:multiple-LF-constraints}; 
$p$ is the proportion of runs and $\mu$ is the average expansion factor $\alpha$ with s.d.}
\label{tab:feasibility-guard-summary}
\begin{center}
\footnotesize
\begin{tabular}{lccc}
\toprule
\multicolumn{4}{c}{\bf (A) \texttt{ROCF-AD/LF}} \\
\addlinespace[0.35em]
\multicolumn{1}{c}{\bf $\vec{\delta}{=}(\delta_{\textrm{DP}}, \delta_{\textrm{EOpp}}, \delta_{\textrm{PEq}}, \delta_{\textrm{PP}}, \delta_{\textrm{FOR}})$ } & \multicolumn{1}{c}{\bf COMPAS} & \multicolumn{1}{c}{\bf ACSIncome ($|\mathcal{A}|{=}2$)} & \multicolumn{1}{c}{\bf ACSIncome ($|\mathcal{A}|{=}5$)} \\
\midrule
$\vec{\delta}=(0.05,0.05,0.05,0.05,\mathrm{None})$ & p=0.00 ; $\mu$=$1.00\pm0.00$ & p=0.00 ; $\mu$=$1.00\pm0.00$ & p=0.00 ; $\mu$=$1.00\pm0.00$ \\
$\vec{\delta}=(\mathrm{None},0.1,\mathrm{None},0.1,0.1)$ & p=0.10 ; $\mu$=$1.01\pm0.02$ & p=0.00 ; $\mu$=$1.00\pm0.00$ & p=0.66 ; $\mu$=$1.04\pm0.05$ \\
\addlinespace[0.35em]
\multicolumn{4}{c}{\bf (B) \texttt{Oracle}} \\
\addlinespace[0.35em]
\multicolumn{1}{c}{\bf $\vec{\delta}{=}(\delta_{\textrm{DP}}, \delta_{\textrm{EOpp}}, \delta_{\textrm{PEq}}, \delta_{\textrm{PP}}, \delta_{\textrm{FOR}})$ } & \multicolumn{1}{c}{\bf COMPAS} & \multicolumn{1}{c}{\bf ACSIncome ($|\mathcal{A}|{=}2$)} & \multicolumn{1}{c}{\bf ACSIncome ($|\mathcal{A}|{=}5$)} \\
\midrule
$\vec{\delta}=(0.05,0.05,0.05,0.05,\mathrm{None})$ & p=0.00 ; $\mu$=$1.00\pm0.00$ & p=0.00 ; $\mu$=$1.00\pm0.00$ & p=0.00 ; $\mu$=$1.00\pm0.00$\\
$\vec{\delta}=(\mathrm{None},0.1,\mathrm{None},0.1,0.1)$ & p=0.08 ; $\mu$=$1.00\pm0.02$ & p=0.00 ; $\mu$=$1.00\pm0.00$ & p=0.64 ; $\mu$=$1.05\pm0.08$ \\
\bottomrule
\end{tabular}
\end{center}
\end{table}

\subsubsection{Computational Runtimes}\label{app:add-exp:runtimes}

\begin{figure}
\begin{center}
\includegraphics[width=0.98\linewidth]{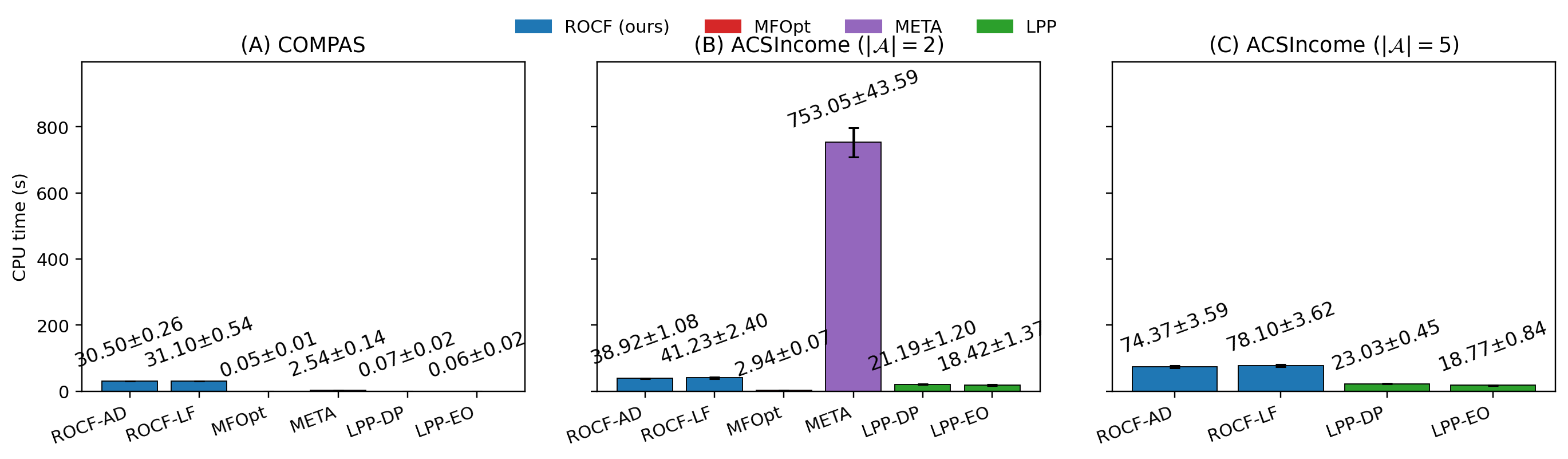}
\end{center}
\caption{ CPU runtime (mean $\pm$ s.e., seconds)
for (A) COMPAS, (B) ACSIncome ($|\mathcal{A}|{=}2$), and (C) ACSIncome ($|\mathcal{A}|{=}5$). As in \S\ref{sec:exp:primary-result} and \S\ref{sec:exp:multiple-protected-attributes}, the disparities  $\delta_{\textrm{DP}},\delta_{\textrm{EOpp}},\delta_{\textrm{PEq}}$ and $\delta_{\textrm{PP}}$ are controlled at level $0.05$ whenever active.}
\label{fig:runtime-compas-acsincome}
\end{figure}

\begin{figure}
\begin{center}
\includegraphics[width=0.98\linewidth]{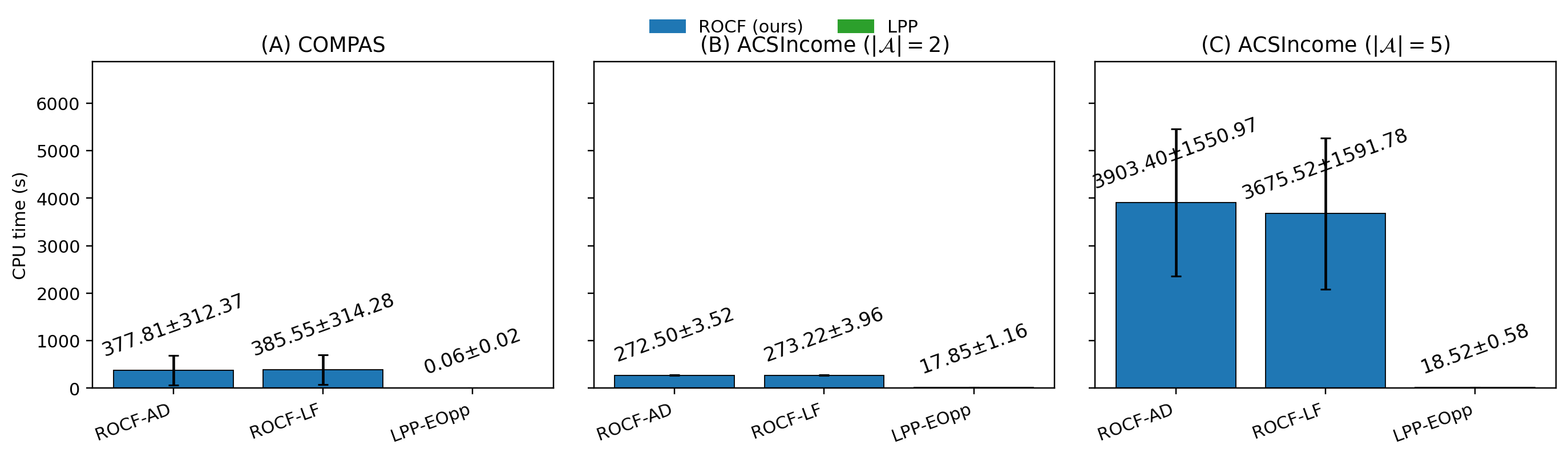}
\end{center}
\caption{ CPU runtime (mean $\pm$ s.e., seconds) 
for (A) COMPAS, (B) ACSIncome ($|\mathcal{A}|{=}2$), and (C) ACSIncome ($|\mathcal{A}|{=}5$). As in \S\ref{sec:exp:multiple-LF-constraints}, the disparities $\delta_{\textrm{EOpp}},\delta_{\textrm{PP}}$ and $\delta_{\textrm{FOR-parity}}$ are controlled at level $0.05$ for our methods \texttt{ROCF-AD/LF}, while \texttt{LPP-EOpp} only controls for equality of opportunity.}
\label{fig:runtime-compas-acsincome-multiple-LF-constraints}
\end{figure}

In this section, we report wall-clock running times of our methods and baselines on the experiments in \S\ref{sec:empirical}. 
Figures \ref{fig:runtime-compas-acsincome} 
and \ref{fig:runtime-compas-acsincome-multiple-LF-constraints}
display the runtimes measuring the average execution time of a single run per seed for the COMPAS and ACSIncome datasets.
All experiments are conducted on a computing cluster equipped with Intel Xeon Platinum 8375C CPUs $@$ 2.90GHz processors. For the COMPAS dataset, each run is allowed to use up to 10 CPU cores with 1 GB of RAM, while for ACSIncome, each run uses a single core with access of up to 32 GB of RAM.

When the nominal levels are well specified and the guard does not trigger, runtimes are modest with small error bars (Figure~\ref{fig:runtime-compas-acsincome} and Figure~\ref{fig:runtime-compas-acsincome-multiple-LF-constraints}(B)).
We observe that this setting occurs often in practice; see Table~\ref{tab:feasibility-guard-summary}.
By contrast, under the multiple LF constraint setting, the feasibility guard is often triggered, so we require a bisection section that adds multiple \texttt{RegionSearch} (Algorithm~\ref{alg:region-search}) solves.
This leads to both larger means and larger standard errors in Figure~\ref{fig:runtime-compas-acsincome-multiple-LF-constraints}(A),(C).

Although \texttt{ROCF-AD/LF} are slower than classic post-processing baselines, the longest observed runtime (on \textsc{ACSIncome} with $ |\mathcal{A}|{=}5$ and $\approx\!1.6$M individuals) is on the order of minutes ($\approx$60 minutes) and remains practical for offline model selection.
More importantly, these additional solves deliver strong empirical results: \texttt{ROCF-AD/LF} attains the nominal disparity levels while achieving nearly optimal accuracy-fairness tradeoffs (cf. \S\ref{sec:empirical}).

\subsubsection{Results on post-processing set}\label{app:add-exp:compas-acsincome-post}

This section reports the performance of our methods along with baselines on the post-processing set. 
By design, our methods \texttt{ROCF-AD/LF} attain either the nominal disparity levels when the feasibility guard (cf. \S\ref{sec:region-search}) is not activated, or
a minimal relaxation of the nominal disparity level when the guard is activated; see  
Tables \ref{tab:compas-acsincome-post} and \ref{tab:compas-acsincome-multiple-LF-constraints-post}.

\begin{table}
\footnotesize
\caption{ Performance on the post-processing set for (A) COMPAS ($|\mathcal{A}|{=}2$), (B) ACSIncome ($|\mathcal{A}|{=}2$), and (C) ACSIncome ($|\mathcal{A}|{=}5$). 
The disparities $\delta_{\textrm{DP}}, \delta_{\textrm{EOpp}}, \delta_{\textrm{PEq}}, \delta_{\textrm{PP}}$ are controlled at level $0.05$ whenever they are active. 
\textbf{Interv.} is the empirical intervention rate on the post-processing set (see Definition~\ref{def:intervention}). 
Cells in green indicate that the fairness constraint is satisfied within two standard deviations at level $0.05$, whereas cells in red indicate violation. 
Entries in the  Oracle rows are shaded in lighter colors to denote that they are not practically feasible baselines. }
\label{tab:compas-acsincome-post} 
\begin{center}
\begin{tabular}{lccccccc}
\toprule
\multicolumn{8}{c}{\bf (A) COMPAS} \\
\addlinespace[0.35em]
\multicolumn{1}{c}{\bf Method} & \multicolumn{1}{c}{\bf Acc} & \multicolumn{1}{c}{\bf DP} & \multicolumn{1}{c}{\bf EOpp} & \multicolumn{1}{c}{\bf PEq} & \multicolumn{1}{c}{\bf PP} & \multicolumn{1}{c}{\bf FOR} & \multicolumn{1}{c}{\bf Interv.} \\
\midrule
Baseline & 0.68 $\pm$ 0.01 & \rcell{0.27 $\pm$ 0.04} & \rcell{0.26 $\pm$ 0.06} & \rcell{0.19 $\pm$ 0.05} & \gcell{0.07 $\pm$ 0.04} & {0.03 $\pm$ 0.02} & 0.00 $\pm$ 0.00 \\
\addlinespace[0.25em]\cdashline{1-8}[0.4pt/1pt]\addlinespace[0.15em]
Oracle & 0.62 $\pm$ 0.02 
& \gcelloracle{0.05 $\pm$ 0.01} 
& \gcelloracle{0.03 $\pm$ 0.01} 
& \gcelloracle{0.05 $\pm$ 0.01} 
& \gcelloracle{0.05 $\pm$ 0.00} 
& {0.15 $\pm$ 0.02} 
& N/A \\
{\bf ROCF-AD} (ours) & 0.62 $\pm$ 0.02 & \gcell{0.05 $\pm$ 0.01} & \gcell{0.03 $\pm$ 0.02} & \gcell{0.05 $\pm$ 0.01} & \gcell{0.05 $\pm$ 0.01} & {0.16 $\pm$ 0.02} & 0.06 $\pm$ 0.02 \\
{\bf ROCF-LF} (ours) & 0.62 $\pm$ 0.02 & \gcell{0.05 $\pm$ 0.01} & \gcell{0.03 $\pm$ 0.02} & \gcell{0.05 $\pm$ 0.02} & \gcell{0.05 $\pm$ 0.02} & {0.16 $\pm$ 0.03} & 0.06 $\pm$ 0.02 \\
\addlinespace[0.25em]\cdashline{1-8}[0.4pt/1pt]\addlinespace[0.15em]
MFOpt & 0.63 $\pm$ 0.01 & \rcell{0.26 $\pm$ 0.04} & \rcell{0.24 $\pm$ 0.05} & \rcell{0.21 $\pm$ 0.04} & \gcell{0.09 $\pm$ 0.03} & {0.07 $\pm$ 0.03} & 0.14 $\pm$ 0.02 \\
META & 0.51 $\pm$ 0.02 & \gcell{0.05 $\pm$ 0.17} & \gcell{0.05 $\pm$ 0.15} & \gcell{0.04 $\pm$ 0.18} & \gcell{0.06 $\pm$ 0.19} & {0.15 $\pm$ 0.04} & 0.00 $\pm$ 0.00 \\
\addlinespace[0.25em]\cdashline{1-8}[0.4pt/1pt]\addlinespace[0.15em]
LPP-DP & 0.67 $\pm$ 0.00 & \gcell{0.05 $\pm$ 0.00} & \gcell{0.04 $\pm$ 0.00} & \gcell{0.03 $\pm$ 0.00} & \rcell{0.16 $\pm$ 0.00} & {0.09 $\pm$ 0.00} & 0.00 $\pm$ 0.00 \\
LPP-EO & 0.67 $\pm$ 0.00 & \rcell{0.10 $\pm$ 0.00} & \rcell{0.09 $\pm$ 0.00} & \gcell{0.03 $\pm$ 0.00} & \rcell{0.14 $\pm$ 0.00} & {0.07 $\pm$ 0.00} & 0.00 $\pm$ 0.00 \\
\bottomrule
\end{tabular}


\begin{tabular}{lccccccc}
\toprule
\multicolumn{8}{c}{\bf (B) ACSIncome ($|\mathcal{A}|{=}2$)} \\
\addlinespace[0.35em]
\multicolumn{1}{c}{\bf Method} & \multicolumn{1}{c}{\bf Acc} & \multicolumn{1}{c}{\bf DP} & \multicolumn{1}{c}{\bf EOpp} & \multicolumn{1}{c}{\bf PEq} & \multicolumn{1}{c}{\bf PP} & \multicolumn{1}{c}{\bf FOR} & \multicolumn{1}{c}{\bf Interv.} \\
\midrule
Baseline & 0.79 $\pm$ 0.01 & \rcell{0.17 $\pm$ 0.03} & \rcell{0.14 $\pm$ 0.02} & \gcell{0.08 $\pm$ 0.03} & \gcell{0.05 $\pm$ 0.01} & {0.06 $\pm$ 0.02} & 0.00 $\pm$ 0.00 \\
\addlinespace[0.25em]\cdashline{1-8}[0.4pt/1pt]\addlinespace[0.15em]
Oracle & 0.75 $\pm$ 0.00 
& \gcelloracle{0.05 $\pm$ 0.00} 
& \gcelloracle{0.05 $\pm$ 0.00} 
& \gcelloracle{0.02 $\pm$ 0.00} 
& \gcelloracle{0.05 $\pm$ 0.00} 
& {0.14 $\pm$ 0.00} 
& N/A \\
{\bf ROCF-AD} (ours) & 0.75 $\pm$ 0.00 & \gcell{0.05 $\pm$ 0.00} & \gcell{0.05 $\pm$ 0.00} & \gcell{0.02 $\pm$ 0.00} & \gcell{0.05 $\pm$ 0.00} & {0.14 $\pm$ 0.00} & 0.02 $\pm$ 0.00 \\
{\bf ROCF-LF} (ours) & 0.75 $\pm$ 0.00 & \gcell{0.05 $\pm$ 0.00} & \gcell{0.05 $\pm$ 0.00} & \gcell{0.02 $\pm$ 0.00} & \gcell{0.05 $\pm$ 0.00} & {0.14 $\pm$ 0.00} & 0.02 $\pm$ 0.00 \\
\addlinespace[0.25em]\cdashline{1-8}[0.4pt/1pt]\addlinespace[0.15em]
MFOpt & 0.74 $\pm$ 0.00 & \rcell{0.20 $\pm$ 0.00} & \rcell{0.09 $\pm$ 0.00} & \rcell{0.15 $\pm$ 0.00} & \rcell{0.06 $\pm$ 0.00} & {0.07 $\pm$ 0.00} & 0.06 $\pm$ 0.00 \\
META & 0.79 ± 0.00 & \rcell{0.11 ± 0.02} & \gcell{0.03 ± 0.02} & \gcell{0.03 ± 0.01} & \rcell{0.11 ± 0.01} & 0.09 ± 0.01 & 0.00 ± 0.00 \\
\addlinespace[0.25em]\cdashline{1-8}[0.4pt/1pt]\addlinespace[0.15em]
LPP-DP & 0.78 $\pm$ 0.00 & \gcell{0.05 $\pm$ 0.00} & \gcell{0.04 $\pm$ 0.00} & \gcell{0.02 $\pm$ 0.00} & \rcell{0.14 $\pm$ 0.00} & {0.12 $\pm$ 0.00} & 0.00 $\pm$ 0.00 \\
LPP-EO & 0.78 $\pm$ 0.00 & \rcell{0.10 $\pm$ 0.00} & \gcell{0.03 $\pm$ 0.00} & \gcell{0.03 $\pm$ 0.00} & \rcell{0.10 $\pm$ 0.00} & {0.09 $\pm$ 0.00} & 0.00 $\pm$ 0.00 \\
\bottomrule
\end{tabular}


\begin{tabular}{lccccccc}
\toprule
\multicolumn{8}{c}{\bf (C) ACSIncome ($|\mathcal{A}|{=}5$)} \\
\addlinespace[0.35em]
\multicolumn{1}{c}{\bf Method} & \multicolumn{1}{c}{\bf Acc} & \multicolumn{1}{c}{\bf DP} & \multicolumn{1}{c}{\bf EOpp} & \multicolumn{1}{c}{\bf PEq} & \multicolumn{1}{c}{\bf PP} & \multicolumn{1}{c}{\bf FOR} & \multicolumn{1}{c}{\bf Interv.} \\
\midrule
Baseline & 0.79 $\pm$ 0.00 & \rcell{0.25 $\pm$ 0.01} & \rcell{0.24 $\pm$ 0.02} & \rcell{0.10 $\pm$ 0.02} & \rcell{0.23 $\pm$ 0.02} & {0.08 $\pm$ 0.01} & 0.00 $\pm$ 0.00 \\
\addlinespace[0.25em]\cdashline{1-8}[0.4pt/1pt]\addlinespace[0.15em]
Oracle & 0.69 $\pm$ 0.01 
& \gcelloracle{0.05 $\pm$ 0.00} 
& \gcelloracle{0.05 $\pm$ 0.00} 
& \gcelloracle{0.03 $\pm$ 0.01} 
& \gcelloracle{0.05 $\pm$ 0.00} 
& {0.23 $\pm$ 0.01} 
& N/A \\
{\bf ROCF-AD} (ours) & 0.69 $\pm$ 0.01 & \gcell{0.05 $\pm$ 0.00} & \gcell{0.05 $\pm$ 0.00} & \gcell{0.03 $\pm$ 0.01} & \gcell{0.05 $\pm$ 0.01} & {0.23 $\pm$ 0.01} & 0.03 $\pm$ 0.01 \\
{\bf ROCF-LF} (ours) & 0.69 $\pm$ 0.01 & \gcell{0.05 $\pm$ 0.00} & \gcell{0.05 $\pm$ 0.00} & \gcell{0.03 $\pm$ 0.01} & \gcell{0.05 $\pm$ 0.00} & {0.23 $\pm$ 0.01} & 0.03 $\pm$ 0.01 \\
\addlinespace[0.25em]\cdashline{1-8}[0.4pt/1pt]\addlinespace[0.15em]
LPP-DP & 0.78 $\pm$ 0.00 & \gcell{0.05 $\pm$ 0.00} & \rcell{0.07 $\pm$ 0.00} & \rcell{0.09 $\pm$ 0.00} & \rcell{0.35 $\pm$ 0.00} & {0.15 $\pm$ 0.00} & 0.00 $\pm$ 0.00 \\
LPP-EO & 0.78 $\pm$ 0.00 & \rcell{0.12 $\pm$ 0.00} & \rcell{0.06 $\pm$ 0.00} & \rcell{0.06 $\pm$ 0.00} & \rcell{0.33 $\pm$ 0.00} & {0.13 $\pm$ 0.00} & 0.00 $\pm$ 0.00 \\
\bottomrule
\end{tabular}
\end{center}
\end{table}

\begin{table}
\footnotesize
\caption{ Performance on the post-processing set for (A) COMPAS ($|\mathcal{A}|{=}2$), (B) ACSIncome ($|\mathcal{A}|{=}2$), and (C) ACSIncome ($|\mathcal{A}|{=}5$). 
The disparities $\delta_{\textrm{EOpp}}, \delta_{\textrm{PP}}$ and $\delta_{\textrm{FOR}}$ are controlled at level $0.10$ whenever they are active. 
\textbf{Interv.} is the empirical intervention rate on the post-processing set (see Definition~\ref{def:intervention}). 
Cells in green indicate that the fairness constraint is satisfied within two standard deviations at level $0.10$, whereas cells in red indicate violation. 
Entries in the Oracle rows are shaded in lighter colors to denote that they are not practically feasible baselines. }
\label{tab:compas-acsincome-multiple-LF-constraints-post}
\begin{center} 
\begin{tabular}{lccccccc}
\toprule
\multicolumn{8}{c}{\bf (A) COMPAS} \\
\addlinespace[0.35em]
\multicolumn{1}{c}{\bf Method} & \multicolumn{1}{c}{\bf Acc} & \multicolumn{1}{c}{\bf DP} & \multicolumn{1}{c}{\bf EOpp} & \multicolumn{1}{c}{\bf PEq} & \multicolumn{1}{c}{\bf PP} & \multicolumn{1}{c}{\bf FOR} & \multicolumn{1}{c}{\bf Interv.} \\
\midrule
Baseline & 0.68 $\pm$ 0.01 & 0.27 $\pm$ 0.04 & \rcell{0.26 $\pm$ 0.06} & 0.19 $\pm$ 0.05 & \gcell{0.07 $\pm$ 0.04} & \gcell{0.03 $\pm$ 0.02} & 0.00 $\pm$ 0.00 \\
\addlinespace[0.25em]\cdashline{1-8}[0.4pt/1pt]\addlinespace[0.15em]
Oracle & 0.65 $\pm$ 0.04 & 0.15 $\pm$ 0.04 & \gcelloracle{0.10 $\pm$ 0.00} & 0.12 $\pm$ 0.08 & \gcelloracle{0.10 $\pm$ 0.01} & \gcelloracle{0.08 $\pm$ 0.02} & N/A \\
\textbf{ROCF-AD} (ours) & 0.65 $\pm$ 0.04 & 0.15 $\pm$ 0.04 & \gcell{0.10 $\pm$ 0.01} & 0.11 $\pm$ 0.09 & \gcell{0.10 $\pm$ 0.01} & \gcell{0.08 $\pm$ 0.03} & 0.01 $\pm$ 0.02 \\
\textbf{ROCF-LF} (ours) & 0.65 $\pm$ 0.04 & 0.15 $\pm$ 0.04 & \gcell{0.10 $\pm$ 0.01} & 0.12 $\pm$ 0.09 & \gcell{0.10 $\pm$ 0.01} & \gcell{0.09 $\pm$ 0.03} & 0.01 $\pm$ 0.02 \\
\addlinespace[0.25em]\cdashline{1-8}[0.4pt/1pt]\addlinespace[0.15em]
LPP-EOpp & 0.67 $\pm$ 0.01 & 0.15 $\pm$ 0.01 & \gcell{0.13 $\pm$ 0.03} & 0.07 $\pm$ 0.02 & \gcell{0.12 $\pm$ 0.03} & \gcell{0.06 $\pm$ 0.03} & 0.00 $\pm$ 0.00 \\
\bottomrule
\end{tabular}


\begin{tabular}{lccccccc}
\toprule
\multicolumn{8}{c}{\bf (B) ACSIncome ($|\mathcal{A}|{=}2$)} \\
\addlinespace[0.35em]
\multicolumn{1}{c}{\bf Method} & \multicolumn{1}{c}{\bf Acc} & \multicolumn{1}{c}{\bf DP} & \multicolumn{1}{c}{\bf EOpp} & \multicolumn{1}{c}{\bf PEq} & \multicolumn{1}{c}{\bf PP} & \multicolumn{1}{c}{\bf FOR} & \multicolumn{1}{c}{\bf Interv.} \\
\midrule
Baseline & 0.79 $\pm$ 0.01 & 0.17 $\pm$ 0.03 & \gcell{0.14 $\pm$ 0.02} & 0.08 $\pm$ 0.03 & \gcell{0.05 $\pm$ 0.01} & \gcell{0.06 $\pm$ 0.02} & 0.00 $\pm$ 0.00 \\
\addlinespace[0.25em]\cdashline{1-8}[0.4pt/1pt]\addlinespace[0.15em]
Oracle & 0.79 $\pm$ 0.00 & 0.16 $\pm$ 0.00 & \gcelloracle{0.10 $\pm$ 0.00} & 0.07 $\pm$ 0.01 & \gcelloracle{0.07 $\pm$ 0.01} & \gcelloracle{0.07 $\pm$ 0.00} & N/A \\
\textbf{ROCF-AD} (ours) & 0.79 $\pm$ 0.00 & 0.16 $\pm$ 0.00 & \gcell{0.10 $\pm$ 0.00} & 0.07 $\pm$ 0.01 & \gcell{0.07 $\pm$ 0.01} & \gcell{0.07 $\pm$ 0.00} & 0.00 $\pm$ 0.00 \\
\textbf{ROCF-LF} (ours) & 0.79 $\pm$ 0.00 & 0.16 $\pm$ 0.00 & \gcell{0.10 $\pm$ 0.00} & 0.07 $\pm$ 0.01 & \gcell{0.07 $\pm$ 0.01} & \gcell{0.07 $\pm$ 0.00} & 0.00 $\pm$ 0.00 \\
\addlinespace[0.25em]\cdashline{1-8}[0.4pt/1pt]\addlinespace[0.15em]
LPP-EOpp & 0.79 $\pm$ 0.01 & 0.14 $\pm$ 0.01 & \gcell{0.09 $\pm$ 0.01} & 0.05 $\pm$ 0.02 & \gcell{0.08 $\pm$ 0.01} & \gcell{0.07 $\pm$ 0.01} & 0.00 $\pm$ 0.00 \\
\bottomrule
\end{tabular}


\begin{tabular}{lccccccc}
\toprule
\multicolumn{8}{c}{\bf (C) ACSIncome ($|\mathcal{A}|{=}5$)} \\
\addlinespace[0.35em]
\multicolumn{1}{c}{\bf Method} & \multicolumn{1}{c}{\bf Acc} & \multicolumn{1}{c}{\bf DP} & \multicolumn{1}{c}{\bf EOpp} & \multicolumn{1}{c}{\bf PEq} & \multicolumn{1}{c}{\bf PP} & \multicolumn{1}{c}{\bf FOR} & \multicolumn{1}{c}{\bf Interv.} \\
\midrule
Baseline & 0.78 $\pm$ 0.01 & 0.25 $\pm$ 0.01 & \rcell{0.24 $\pm$ 0.03} & 0.10 $\pm$ 0.02 & \rcell{0.23 $\pm$ 0.03} & \gcell{0.08 $\pm$ 0.01} & 0.00 $\pm$ 0.00 \\
\addlinespace[0.25em]\cdashline{1-8}[0.4pt/1pt]\addlinespace[0.15em]
Oracle & 0.57 ± 0.02 & 0.48 ± 0.03 & \gcelloracle{0.10 ± 0.01} & 0.57 ± 0.05 & \gcelloracle{0.10 ± 0.01} & \gcelloracle{0.10 ± 0.01} & N/A \\
\textbf{ROCF-AD} (ours) & 0.57 ± 0.02 & 0.48 ± 0.03 & \gcell{0.10 ± 0.01} & 0.57 ± 0.05 & \gcell{0.10 ± 0.01} & \gcell{0.11 ± 0.02} & 0.06 ± 0.01 \\
\textbf{ROCF-LF} (ours) & 0.56 ± 0.02 & 0.49 ± 0.03 & \gcell{0.10 ± 0.01} & 0.58 ± 0.04 & \gcell{0.10 ± 0.01} & \gcell{0.10 ± 0.01} & 0.05 ± 0.01 \\
\addlinespace[0.25em]\cdashline{1-8}[0.4pt/1pt]\addlinespace[0.15em]
LPP-EOpp & 0.79 $\pm$ 0.00 & 0.15 $\pm$ 0.01 & \gcell{0.11 $\pm$ 0.01} & 0.06 $\pm$ 0.01 & \rcell{0.31 $\pm$ 0.02} & \gcell{0.12 $\pm$ 0.01} & 0.00 $\pm$ 0.00 \\
\bottomrule
\end{tabular}
\end{center}
\end{table}

\end{document}